\documentclass{article}
\usepackage{PRIMEarxiv}
\usepackage[utf8]{inputenc} 
\usepackage[T1]{fontenc}    
\usepackage[utf8]{vietnam}
\usepackage[english]{babel}
\usepackage{tipa}
\usepackage{ragged2e}

\usepackage{hyperref}       
\usepackage{url}            
\usepackage{booktabs}       
\usepackage{amsfonts}       
\usepackage{amsmath}
\usepackage{nicefrac}       
\usepackage{microtype}      
\usepackage{lipsum}
\usepackage{graphicx}
\graphicspath{{figure/}}     
\usepackage{caption}
\usepackage{subcaption}
\usepackage{multirow}
\usepackage{booktabs}

\usepackage[final]{pdfpages}
\usepackage{tabularx}
\usepackage{tkz-tab}
\usepackage{framed,multicol}
\usepackage[framemethod=tikz]{mdframed}
\usepackage{xcolor}

\usepackage{wasysym}
\usepackage{pgfplots}
\usepackage{tikz}
\usepackage{pgfplotstable}
\usepackage{filecontents}

\usepgfplotslibrary{colorbrewer}
\pgfplotsset{compat = 1.15, cycle list/Set1-8} 
\usetikzlibrary{pgfplots.statistics, pgfplots.colorbrewer} 

\pgfplotsset{compat=1.3}
\usetikzlibrary{positioning, fit, calc}  
\tikzset{block/.style={draw, thick, text width=2cm ,minimum height=1.3cm, align=center},   
	line/.style={-latex}     
} 

\tikzset{blocktext/.style={draw, thick, text width=5.2cm ,minimum height=1.3cm, align=center},   
	line/.style={-latex}     
}

\tikzset{font=\footnotesize}

\title{Investigating the Effectiveness of ChatGPT in Mathematical Reasoning and Problem Solving: Evidence from the Vietnamese National High School Graduation Examination}

\author{ \href{https://orcid.org/0000-0001-5415-7538}{\includegraphics[scale=0.06]{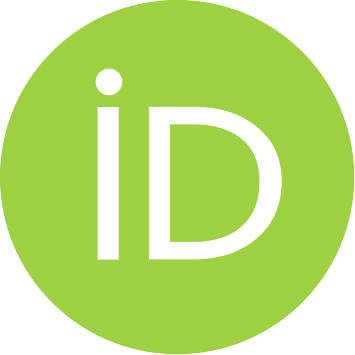}\hspace{1mm}Xuan-Quy Dao} \\
	School of Engineering\\
	Eastern International University\\
	Binh Duong, Vietnam \\
	\texttt{quy.dao@eiu.edu.vn} \\
	\And
	\href{https://orcid.org/0000-0001-7431-0157}{\includegraphics[scale=0.06]{orcid.pdf}\hspace{1mm}Ngoc-Bich Le} \\
	School of Biomedical Engineering \\
	International University,
	VNUHCM City\\
	Ho Chi Minh City, Vietnam \\
	\texttt{lnbich@hcmiu.edu.vn} \\
}

\hypersetup{
pdftitle={Evaluation of ChatGPT's Capabilities in Mathematics Test of the Vietnamese High School Graduation Examination},
pdfsubject={ChatGPT},
pdfauthor={Xuan-Quy DAO},
pdfkeywords={First keyword, Second keyword, More},
}

\begin{document}
\maketitle

\begin{abstract}

This study offers a complete analysis of ChatGPT's mathematics abilities in responding to multiple-choice questions for the Vietnamese National High School Graduation Examination (VNHSGE) on a range of subjects and difficulty levels. The dataset included 250 questions divided into four levels: knowledge (K), comprehension (C), application (A), and high application (H), and it included ten themes that covered diverse mathematical concepts. The outcomes demonstrate that ChatGPT's performance varies depending on the difficulty level and subject. It performed best on questions at Level (K), with an accuracy rate of $83\%$; but, as the difficulty level rose, it scored poorly, with an accuracy rate of $10\%$. The study has also shown that ChatGPT significantly succeeds in providing responses to questions on subjects including exponential and logarithmic functions, geometric progression, and arithmetic progression. The study found that ChatGPT had difficulty correctly answering questions on topics including derivatives and applications, spatial geometry, and Oxyz spatial calculus. Additionally, this study contrasted ChatGPT outcomes with Vietnamese students in VNHSGE and in other math competitions. ChatGPT dominated in the SAT Math competition with a success rate of $70\%$, followed by VNHSGE mathematics ($58.8\%)$. However, its success rates were lower on other exams, such as AP Statistics, the GRE Quantitative, AMC 10, AMC 12, and AP Calculus BC. These results suggest that ChatGPT has the potential to be an effective teaching tool for mathematics, but more work is needed to enhance its handling of graphical data and address the challenges presented by questions that are getting more challenging.

\end{abstract}


\section{Introduction}

In recent years, artificial intelligence (AI) has drawn a lot of interest and been extensively discussed. AI represents a creative and imaginative advancement in many fields, including mathematics instruction. The current work analyzes a number of studies that looked into the application of AI in a number of contexts, including medical~\cite{he2019practical}, education~\cite{chen2020artificial},~\cite{cope2021artificial},~\cite{Dao2021},~\cite{Nguyen2021} and pandemics~\cite{vaishya2020artificial}. The role of educators should not be replaced by AI in the educational process; rather, AI should be used to enhance it~\cite{popenici2017exploring}. The implementation of AI in education faces a variety of challenges despite the potential benefits.

In order to improve student learning outcomes and get around obstacles like a shortage of qualified teachers and resources~\cite{zhang2021ai},~\cite{zawacki2019systematic}, using AI in education is becoming more popular~\cite{zafari2022artificial},~\cite{pedro2019artificial},\cite{ahmad2021artificial},~\cite{paek2021analysis},~\cite{zheng2021effectiveness}. According to research, AI is crucial for guaranteeing sustainable societal growth and can boost student accomplishment. Despite the fact that literature evaluations have been undertaken on the use of AI in education across a variety of subjects, little is known about how AI especially affects mathematics education, including its nature, target grade levels, and study methodologies. Achievement in mathematics is important for kids' academic progress, future employment prospects, and social growth, and it is connected to civil rights issues~\cite{gamoran2000algebra},~\cite{moses2002radical}. Therefore, preparing students with math skills and knowledge is crucial for adapting to a society that is changing quickly and ensuring sustainable development. A comprehensive literature review was undertaken by bin Mohamed et al.~\cite{bin2022artificial} to provide an overview of AI in mathematics education for students at all levels of education, one of the few studies on the effects of AI on mathematics education. This review contributes to the discussion about enhancing teaching and learning in mathematics education through the use of AI. In a different study, Hwang~\cite{hwang2022examining} used 21 empirical studies with 30 independent samples to conduct a meta-analysis to assess the overall impact of AI on elementary children' mathematical achievement. The results of the study revealed that AI had a negligible impact on primary kids' mathematical proficiency. The results showed that grade level and topic of mathematics learning variables considerably reduced the impact of AI on mathematical achievement. Other moderator variables' effects, however, were found to be insignificant. Based on the findings, this study offers both practical and theoretical insights that can help guide the appropriate application of AI in the teaching of mathematics to elementary school children. It is evident that additional meta-analysis is required to determine whether AI offers novel opportunities for mathematics learning~\cite{ahmad2021artificial},~\cite{zheng2021effectiveness}. Studies examining how moderating variables affect the connection between them are also necessary.

The area of education could undergo a revolution owing to recent advancements in natural language processing (NLP), which have led to the development of increasingly complex language models like GPT-3. Due to its capacity to produce natural language answers to a variety of questions, ChatGPT, a large language model based on the GPT architecture, has attracted a great deal of interest in the educational community. In recent years, there has been an increase in interest in using chatbots, particularly ChatGPT, in education. Several research have investigated the possible advantages, issues, and difficulties of this practice. Halaweh~\cite{halaweh2023chatgpt} addressed educators' worries about the adoption of ChatGPT into educational contexts, arguing for its inclusion and offering guidelines for safe implementation. In a research on the potential effects of ChatGPT on education, Zhai~\cite{Zhai2023} recommended changing instructional objectives to emphasize students' creativity and critical thinking. In their discussion of the possible advantages and difficulties of employing large language models in educational contexts, Kasneci et al.~\cite{kasneci2023chatgpt} placed emphasis on the requirement for competences and literacies to comprehend the technology and its constraints.

The effectiveness of ChatGPT in assessments has also been examined in studies. (Kortemeyer, 2023) discovered that ChatGPT displayed several misconceptions and mistakes typical of a beginner learner yet would only about pass a calculus-based physics course. Katz et al.~\cite{katz2023gpt} conducted an experimental evaluation of GPT-4's zero-shot performance on the complete Uniform Bar Examination (UBE), demonstrating that it performed better than human test-takers and previous models on the Multistate Bar Examination (MBE), which is a multiple-choice test. Gilson et al.~\cite{gilson2023does} assessed ChatGPT's performance on multiple-choice questions related to the USMLE Step 1 and Step 2 tests and discovered that its performance is comparable to a third-year medical student. These studies show the potential of chatbots to enhance education and legal services, but they also raise questions about their accuracy and dependability in assessments.

Through the simulation of various use cases, Frieder et al.~\cite{frieder2023mathematical} conducted a study to evaluate the mathematical proficiency of ChatGPT and determine its potential as a helpful assistant to professional mathematicians. The outcomes revealed that ChatGPT participants' mathematical skills were significantly worse to those of the typical mathematics graduate student. However, it is critical to also assess ChatGPT's mathematical prowess at lower levels, such as high school. This evaluation would shed light on ChatGPT's capacity to support teachers and students in this level of mathematics learning.

NLP has received a lot of attention recently as a vital study area. Chatbots, one of its implementations, have drawn attention for its capacity to mimic human interactions. While current research highlights the potential of chatbots to support students' learning in a variety of educational settings, their effectiveness in completing particular subjects, like mathematics, in high-stakes exams has received little attention. By evaluating ChatGPT's ability to complete mathematical challenges and pass the VNHSGE exam, this study aims to fill this knowledge gap in the literature. This will be achieved by contrasting ChatGPT's performance in our test with that of earlier assessments made by the OpenAI team~\cite{OpenAI_gpt_4_report}. This study intends to advance knowledge of the benefits of utilizing cutting-edge technology in education to enhance student results by studying the efficiency of AI-powered chatbots in assisting students in high-stakes tests. The results of this study may be especially helpful to educators and policymakers who want to use AI to enhance learning outcomes.

In this article, we concentrate on examining ChatGPT's capability for resolving mathematical issues within the framework of the VNHSGE exam. The Vietnamese educational system places a high value on mathematics, which is frequently seen as a key predictor of student achievement. The promise of AI-powered tools for enhancing mathematics education can therefore be shown by analyzing ChatGPT's mathematical capabilities in the context of the VNHSGE mathematics dataset~\cite{dao2023vnhsge}. Our work seeks to evaluate ChatGPT's performance on mathematical inquiries in the VNHSGE exam critically and explore the prospects of deploying AI-powered tools to assist enhance mathematics teaching.

\section{Objectives and Methodology}
\label{sec:methods}

\subsection{Objectives}

This study aims to offer a thorough analysis of ChatGPT's mathematical skills in relation to the mathematics evaluation for the VNHSGE exam. We seek to shed light on the possibilities of AI tools for educational support and investigate their role in changing the educational landscape by evaluating ChatGPT's performance in these areas. This study also attempts to illustrate ChatGPT's shortcomings when dealing with questions that differ from those present in the VNHSGE exam in terms of both structure and level of difficulty.

\subsection{Scope and Limitation}

By analyzing ChatGPT's responses to questions from the VNHSGE exam that involve mathematics, this study seeks to assess ChatGPT's mathematical capabilities. Our objective is to assess how well ChatGPT responds to these questions and to provide details on ChatGPT's potential in the context of Vietnamese education.

It's important to remember that our evaluations are restricted to the unique the VNHSGE exam structure. The results of ChatGPT are incapable of being extrapolated to tests with other numbers or difficulty levels. This restriction highlights the need for caution when extrapolating from our results and making generalizations regarding ChatGPT's potential uses in educational contexts outside the scope of this study.

\subsection{Methods}

In this study, we evaluated the capability of the ChatGPT model to answer mathematical problems in the VNHSGE mathematics dataset~\cite{dao2023vnhsge}. Using a sequence-to-sequence methodology, the model was developed using a dataset of math problems after being trained on a sizable corpus of text. The mathematical problem was the model's input, and the solution was its output. We compared the produced answers from ChatGPT with the accurate responses given in the exam papers in order to evaluate its performance.

We created a detailed process with many phases to carry out this examination. In the beginning, we gathered information from official test papers made available by the Vietnamese Ministry of Education and Training. We chose these questions as an accurate representation of the actual exam because they were all taken from high school mathematics exams.

The data needs to be formatted in a way that ChatGPT could interpret afterward. The exam questions contained mathematical equations and symbols, which we transformed into LaTeX format to display in a uniform manner. The exam questions were then transformed from their LaTeX format into JSON (JavaScript Object Notation), a lightweight data transfer standard that is frequently used in web applications.

We were able to give the questions to the pre-trained ChatGPT model and get its generated answers after formatting the data in a way that ChatGPT could understand. Finally, we determined ChatGPT's performance score by comparing the generated answers to the accurate responses provided by the exam papers.

Overall, this methodology allowed us to thoroughly evaluate ChatGPT's capacity to answer mathematical problems in the VNHSGE exam. By outlining the specific procedures, we took, we intend to offer a framework for future research examining the efficiency of chatbots powered by AI in assisting students in demanding exams.

\section{Dataset}
\label{sec:dataset}

The VNHSGE mathematics test dataset for the academic years 2019–2023 was used in this investigation. 250 multiple-choice math questions covering a range of subjects, such as algebra, geometry, and calculus, make up the dataset. Based on Bloom's Taxonomy, these questions were divided into four difficulty levels: K (knowledge), C (comprehension), A (application), and H (high application). The Vietnamese Ministry of Education and Training publicly released the dataset, which is frequently used to evaluate students' mathematical aptitude.

\subsection{Question Levels}

Different levels of competence in comprehending and using mathematical concepts are necessary for solving mathematical problems. The dataset includes a range of levels of difficulty, from K-based questions that evaluate fundamental understanding to high-application questions that assess the capacity to analyze and synthesize information in order to solve complex problems. This allows for a thorough evaluation of ChatGPT's mathematical problem-solving abilities. Based on the sort of cognitive activity and verbs used in responding to the questions, the four levels of complexity—K, C, A and H—were established. We can learn more about ChatGPT's strengths and drawbacks when we evaluate its performance on a range of mathematical problems of varying degrees of difficulty.

\subsection{Question Topics}

The dataset provides a thorough assessment of ChatGPT participants' mathematical knowledge and abilities by encompassing a wide range of mathematical topics. M11A: Combinations and Probability; M11B: Number Series (Arithmetic progression, Geometric progression); M11C: Spatial Geometry; M12A: Derivatives and Applications; M12B: Exponential and Logarithmic Functions; M12C: Primitives and Integrals; M12D: Complex Numbers; M12E: Polyhedrons; M12F: Rotating Circle Block; and M12G: Oxyz Spatial Calculus. These topics were included to ensure a thorough evaluation of the ChatGPT's mathematical abilities by testing its understanding, application, analysis, and evaluation of mathematical concepts and principles. Researchers can learn about ChatGPT's strengths and limitations and identify opportunities for development by analyzing how well it performs across all of these issues. 

\subsection{Knowledge matrix}

A key element of assessment systems that gives a thorough breakdown of the criteria and content to be evaluated is the question matrix. To create and compile questions for various tests and examinations, this technical design was deployed. It acts as a reference for test designers in choosing appropriate questions that appropriately reflect the educational and learning objectives of the assessment system. By ensuring that the test questions assess the desired knowledge, skills, and abilities of the examiners and that they are aligned with the learning outcomes, the question matrix aids in assuring the validity, reliability, and fairness of the assessment. As a result, the question matrix is an essential tool for creating high-quality tests that accurately assess student achievement and guide educational decisions.

A knowledge matrix, which classifies each question according to its specific level and topic, can effectively depict the structure and substance of an exam. Administrators of exams and educators can gain a lot from employing a knowledge matrix since it can be used to determine where students' knowledge is strong and weak and to build focused interventions to boost performance. Additionally, the knowledge matrix makes sure that the exam covers a wide range of subjects and levels of difficulty, providing a thorough evaluation of student's knowledge and abilities. The usage of a knowledge matrix ensures that exam results accurately reflect students' abilities and accomplishments by increasing the validity and reliability of exam scores.

The knowledge matrix for the VNHSGE exam in Mathematics for the years 2019-2023 is displayed in Table~\ref{tabl:knowledge_matrix_2019_2023}. We have a distribution of questions based on the topics and degree of difficulty. We can identify a specified number of question levels pertinent to the issue based on the distribution. The distribution of questions by level is shown in Figure 1 as follows: knowlegde 103 (41\%), comprehension 77 (31\%), application 41 (16\%), and high application 29 (12\%). M11A -10 (4\%), M11B - 5 (2\%), M12C - 8 (3\%), M12A - 57 (23\%), M12B - 39 (16\%), M12C - 33 (13\%), M12D - $26(10 \%)$, M12E - $17(7 \%)$, M12F - $14(6 \%)$, and M12G - $41(16 \%)$ are the breakdown of questions by type. Generally, the knowledge matrix offers a thorough overview of the exam's structure and content, making it possible to assess and enhance students' mathematical understanding and problem-solving skills. The exam framework does not have a uniform allocation of questions. There are some topics and problems that just call for knowledge and comprehension, not a high-level application. A majority of the questions-roughly $70 \%$-are focused on knowledge and comprehension. In addition, only $10 \%$ of the questions concentrate on information from the 11th grade, while $90 \%$ are at the 12th grade level. Questions on subjects like M12A, M12B, M12G, and M12C are plentiful. It should be emphasized, nonetheless, that the questions in topic M11B only call for a certain level of expertise.

\begin{figure*}[h]
	\begin{center}
		\includegraphics[width=\textwidth]{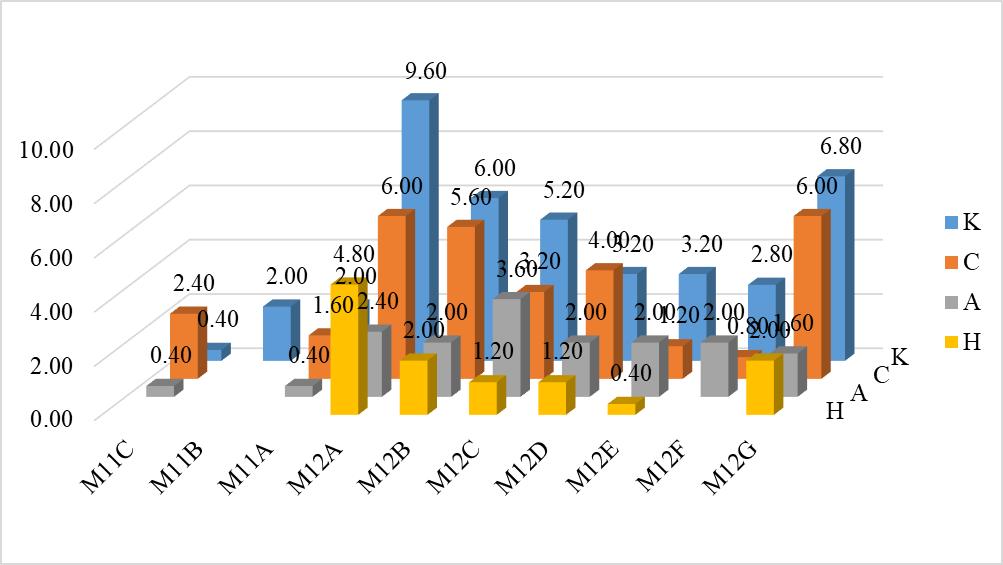}
	\end{center}
	\caption{Distribution of the number of questions by levels and topics in percentage.}
	\label{fig:knowledge_matrix}
\end{figure*}

The distribution of question levels and topics as a percentage is shown in Figure~\ref{fig:knowledge_matrix}. The topic M12A, which comprises 23$\%$ of the total questions, is distributed as follows: 9.60$\%$ at the K level, 6.00$\%$ at the C level, 2.40$\%$ at the A level, and 4.80$\%$ at the H level. We may analyze the performance of the student or ChatGPT specifically by level and topic based on the thorough distribution by level and topic. A comprehensive grasp of the distribution of questions across various levels and topics is made possible by this graphic portrayal. Insights into the areas where test takers are anticipated to perform well and those that could need more improvement can be obtained by examining Figure~\ref{fig:knowledge_matrix}. It offers useful data that teachers and curriculum designers may use to better understand the strengths and weaknesses of their students and the efficiency of their instructional strategies. Overall, Table~\ref{tabl:knowledge_matrix_2019_2023} and Figure~\ref{fig:knowledge_matrix} together give a thorough breakdown of the distribution of the questions and are an effective tool for educational study and practice.

\begingroup
\renewcommand{\arraystretch}{1.5} 
\begin{center}
\begin{table}[ht!]
	\caption{Knowledge matrix in 2019-2023}
	\label{tabl:knowledge_matrix_2019_2023}
	\resizebox{\textwidth}{!}{%
	\begin{tabular}{c|c|c|c|c|c|c|c|c|c|c|cc}
		\cline{2-13}
		\multicolumn{1}{l|}{}                                 & \textbf{M11C} & \textbf{M11B} & \textbf{M11A} & \textbf{M12A} & \textbf{M12B} & \textbf{M12C} & \textbf{M12D} & \textbf{M12E} & \textbf{M12F} & \textbf{M12G} & \multicolumn{2}{c|}{\textbf{LEVEL}}                  \\ \hline
		\multicolumn{1}{|c|}{\textbf{K}}                      & 1             & 5             & 5             & 24            & 15            & 13            & 8             & 8             & 7             & 17            & \multicolumn{1}{c|}{103} & \multicolumn{1}{c|}{41$\%$}  \\ \hline
		\multicolumn{1}{|c|}{\textbf{C}}                      & 6             &               & 4             & 15            & 14            & 8             & 10            & 3             & 2             & 15            & \multicolumn{1}{c|}{77}  & \multicolumn{1}{c|}{31$\%$}  \\ \hline
		\multicolumn{1}{|c|}{\textbf{A}}                      & 1             &               & 1             & 6             & 5             & 9             & 5             & 5             & 5             & 4             & \multicolumn{1}{c|}{41}  & \multicolumn{1}{c|}{16$\%$}  \\ \hline
		\multicolumn{1}{|c|}{\textbf{H}}                      &               &               &               & 12            & 5             & 3             & 3             & 1             &               & 5             & \multicolumn{1}{c|}{29}  & \multicolumn{1}{c|}{12$\%$}  \\ \hline
		\multicolumn{1}{|c|}{\multirow{2}{*}{\textbf{TOPIC}}} & 8             & 5             & 10            & 57            & 39            & 33            & 26            & 17            & 14            & 41            & \multicolumn{1}{c|}{250} &                           \\ \cline{2-13}		\multicolumn{1}{|c|}{}                                & 3$\%$            & 2$\%$            & 4$\%$            & 23$\%$           & 16$\%$           & 13$\%$           & 10$\%$           & 7$\%$            & 6$\%$            & 16$\%$           & \multicolumn{1}{c|}{}    & \multicolumn{1}{c|}{100$\%$} \\ \cline{1-11} \cline{13-13} 
	\end{tabular}		
	}
\end{table}
\end{center}
\endgroup 

\subsection{Prompt and Answer}

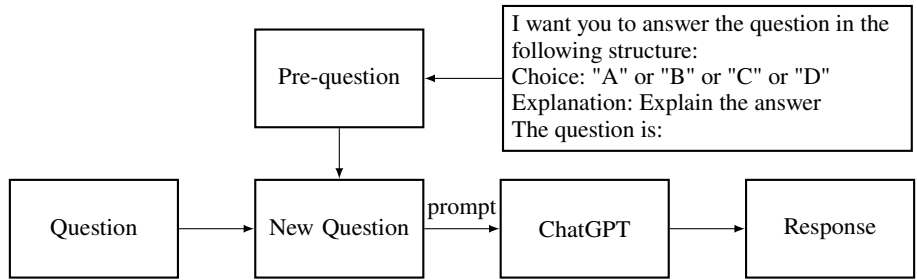
\begin{figure}[ht!]
	\begin{center}
		\begin{tikzpicture}  
			\node[block] (a) {Question};  
			\node[block,right=of a] (b) {New Question};   
			\node[block,right=of b] (c) {ChatGPT};  
			\node[block,right=of c] (d) {Response}; 
			\node[blocktext,align=flush left] (f) at ([yshift=2cm]$(c)!0.5!(d)$) {I want you to answer the question in the following structure:  \\ 
				Choice: "A" or "B" or "C" or "D"  \\
				Explanation: Explain the answer \\
				The question is: {}
			};   
			\node[block] (h) at ([yshift=2cm]$(b)!1.0!(b)$) {Pre-question}; 
			\draw[line] (a)-- (b);  
			\draw[line] (b)-- (c) node [midway, above, sloped] (TextNode) {prompt};  
			\draw[line] (c)-- (d);  
			\draw[line] (h)-- (b);  
			\draw[line] (f)-- (h);  
		\end{tikzpicture} 
	\end{center}
	\caption{Formatted question and ChatGPT response.}
	\label{fig:chatbot_response}
\end{figure}

When asking questions to ChatGPT, we can receive answers in different formats. However, to make the process of handling results easier and ensure consistency, we kindly ask ChatGPT to provide replies in a specific structure. Figure~\ref{fig:chatbot_response} and Table~\ref{tabl:example_chatbot_response} demonstrate an example of the required structure for ChatGPT responses. This table demonstrates the adaptability and versatility of the model by giving instances of how ChatGPT can respond to different cues in various formats. When we receive automatic responses, we utilize Word format on https://chat.openai.com/ but "OpenAI API" uses Json format. The table is divided into three columns: the first column reveals the prompt's format; the second column displays the prompt itself; and the third column provides the response that ChatGPT created. This table demonstrates the adaptability and versatility of the model by giving instances of how ChatGPT can respond to different prompts in various formats. When we receive automatic responses, we utilize Word format on https://chat.openai.com/ but "OpenAI API" uses Json format. The table shows how ChatGPT can provide responses to prompts in many formats, which is a useful feature for many applications. 

\begin{table*}[h!]
	
	\caption{An example of prompt and response.}
	\label{tabl:example_chatbot_response}
	
	\begingroup
	\renewcommand{\arraystretch}{2} 
	
	\textbf{Question (Word format)}:
	
	\begin{tabular}{|l|l|l|l|l|l|}
		\hline
		ID & IQ & Q   & C & IA & E           \\ \hline
		1  &  & \begin{tabular}[c]{@{}l@{}}1) The volume of a cube with edge 2a is:\\ A. 8a\textasciicircum3 \\ B. 2a\textasciicircum3. \\ C. a\textasciicircum3 \\ D. 6a\textasciicircum3.\end{tabular} & A      &        & \begin{tabular}[c]{@{}l@{}}The volume of a cube \\ with edge 2a is: \\ V=(2a)\textasciicircum3=8a\textasciicircum3.\end{tabular} \\ \hline
	\end{tabular}
	\\
	
	\textbf{Question (JSON format)}:
	\{
	"ID": "Q1",
	"IQ": " ",
	"Q": "1) The volume of a cube with edge 2a is:\textbackslash{}nA. 8a\textasciicircum3.\textbackslash{}t\textbackslash{}nB. 2a\textasciicircum3.\textbackslash{}t\textbackslash{}nC. a\textasciicircum3.\textbackslash{}t\textbackslash{}nD. 6a\textasciicircum3.",
	"C": "A",
	"IA": " ",
	"E": "The volume of a cube with edge 2a is: V=(2a)\textasciicircum3=8a\textasciicircum3.",
	\}
	
	\textbf{Pre-question (JSON format)}: "I want you to answer the question in the following structure: " 
	"\textbackslash{}nChoice: "A" or "B" or "C" or "D" " 
	"\textbackslash{}nExplanation: Explain the answer" 
	"\textbackslash{}nThe question is: \{\}"

	\textbf{New Question (Prompt)}:
	I want you to answer the question in the following structure:    
	
	Choice: "A" or "B" or "C" or "D"
	
	Explanation: Explain the answer
	
	The question is: 1) The volume of a cube with edge 2a is:
	A. 8a\textasciicircum3. B. 2a\textasciicircum3. C. a\textasciicircum3. D. 6a\textasciicircum3."
	
	\textbf{Response (JSON format)}:
	
	\{
	"ID": "1",
	"IQ": " ",
	"Q": "1) The volume of a cube with edge 2a is:\textbackslash{}nA. 8a\textasciicircum3.\textbackslash{}t\textbackslash{}nB. 2a\textasciicircum3.\textbackslash{}t\textbackslash{}nC. a\textasciicircum3.\textbackslash{}t\textbackslash{}nD. 6a\textasciicircum3.",
	"C": "A",
	"IA": " ",
	"E": "The volume of a cube with edge 2a is: V=(2a)\textasciicircum3=8a\textasciicircum3.",
	"CC": "A",
	"CE": "The formula for the volume of a cube is V = s\textasciicircum3, where s is the length of one of its sides. Therefore, the volume of the cube with a side length of 2a is: V = (2a)\textasciicircum3 = 8a\textasciicircum3",
	\}
	
	\textbf{Response (Word format)}:
	
	\begin{tabular}{|l|l|l|l|l|l|l|l|}
		\hline
		ID & IQ & Q                                                                                                                                       & C & IA & E                                                                                                    & CC & CE                                                                                                                                                                                                                                      \\ \hline
		1  &    & \begin{tabular}[c]{@{}l@{}}1) The volume of a cube \\ with edge 2a is:\\ A. 8a\textasciicircum3 \\ B. 2a\textasciicircum3. \\ C. a\textasciicircum3 \\ D. 6a\textasciicircum3.\end{tabular} & A &    & \begin{tabular}[c]{@{}l@{}}The volume of a cube \\ with edge 2a is: \\ V=(2a)\textasciicircum3=8a\textasciicircum3.\end{tabular} & A  & \begin{tabular}[c]{@{}l@{}}The formula for the volume of \\ a cube is V = s\textasciicircum3, where s is \\ the length of one of its sides. \\ Therefore, the volume of \\ the cube with a side length \\ of 2a is: V = (2a)\textasciicircum3 = 8a\textasciicircum3\end{tabular} \\ \hline
	\end{tabular}
	
	\endgroup
	
\end{table*}

\section{Results}

The VNHSGE dataset's mathematics exam is intended to evaluate ChatGPT's mathematical knowledge and problem-solving skills. The test consists of 250 questions in the VNHSGE mathematics dataset~\cite{dao2023vnhsge}, divided into ten topics (M11A, M11B, M11C, M12A-M12G) and four degrees of complexity (knowledge, comprehension, application, and high application). The exam aims to provide a thorough assessment of the mathematical knowledge and abilities of ChatGPT candidates by evaluating a wide range of topics. The questions are made to test ChatGPT's understanding, application, evaluation, and analysis of mathematical concepts and principles, ensuring a thorough evaluation of its mathematical skills. This rigorous assessment makes sure that ChatGPT's math-solving abilities are accurately measured and can be used to guide future NLP advances.

\subsection{ChatGPT score}

The results of the mathematics test taken by ChatGPT from 2019 to 2023 are shown in Table~\ref{tabl:chatgpt result}~\cite{dao2023vnhsge}, together with the number of right answers and corresponding score for each year. A score of 5 represents an average performance on a scale from 0 to 10. These outcomes show that ChatGPT performed better than average on the math test. The ChatGPT ranges from 0 to 7 points. This outcome can be attributed to ChatGPT's propensity to accurately respond to a significant portion of questions at the knowledge and comprehension levels, which make up $70\%$ of the total questions. The middle-range ChatGPT score is clear from the fact that only a small number of questions at both the application and high application levels were correctly answered. Further clarification on this point will be provided in the upcoming sections.

\begingroup
\renewcommand{\arraystretch}{1.5} 
\begin{table}[ht!]
	\caption{ChatGPT's performance in 2019-2023}
	\label{tabl:chatgpt result}
	\begin{center}
		\begin{tabular}{ |c|c|c| } 
			\hline
			{Year} & ChatGPT's Performance & ChatGPT's Score  \\ 
			\hline
			{2023} & 27/50                       & 5.4  \\ 
			{2022} & 31/50                       & 6.2  \\ 
			{2021} & 30/50                       & 6 \\
			{2020} & 33/50                       & 6.6  \\
			{2019} & 26/50                       & 5.2 \\
			\hline
			{Average} & 147/250                  & 5.88 \\
			\hline
		\end{tabular}
	\end{center}
\end{table}
\endgroup 

\subsection{ChatGPT’s performance in order question }

\begin{figure*}[h]
	\begin{center}
		\begin{tikzpicture}
			\begin{axis}
				[
				ylabel={Accuracy},
				legend style={at={(0.5,-0.25)}, 	
					anchor=north,legend columns=-1}, 
				symbolic x coords={
					1,
					2,
					3,
					4,
					5,
					6,
					7,
					8,
					9,
					10,
					11,
					12,
					13,
					14,
					15,
					16,
					17,
					18,
					19,
					20,
					21,
					22,
					23,
					24,
					25,
					26,
					27,
					28,
					29,
					30,
					31,
					32,
					33,
					34,
					35,
					36,
					37,
					38,
					39,
					40,
					41,
					42,
					43,
					44,
					45,
					46,
					47,
					48,
					49,
					50,
				},
				xtick=data,
				x tick label style={rotate=90,anchor=east},
				ymin=0,
				xmin=1,
				xmax=50,
				width=\textwidth, 
				height=3cm, 
				width=16cm,
				axis x line*=bottom, axis y line*=left
				]
				\addplot[color=black,mark=*,semithick, mark options={solid,}]
				coordinates{
					(1,100)
					(2,100)
					(3,80)
					(4,40)
					(5,80)
					(6,100)
					(7,60)
					(8,100)
					(9,80)
					(10,100)
					(11,100)
					(12,100)
					(13,100)
					(14,80)
					(15,60)
					(16,60)
					(17,100)
					(18,40)
					(19,80)
					(20,80)
					(21,80)
					(22,60)
					(23,60)
					(24,80)
					(25,60)
					(26,60)
					(27,60)
					(28,80)
					(29,60)
					(30,60)
					(31,40)
					(32,40)
					(33,80)
					(34,80)
					(35,80)
					(36,20)
					(37,60)
					(38,60)
					(39,20)
					(40,20)
					(41,20)
					(42,20)
					(43,20)
					(44,0)
					(45,40)
					(46,0)
					(47,0)
					(48,0)
					(49,40)
					(50,0)
				};
				
			\end{axis}
		\end{tikzpicture}
	\end{center}
	\caption{ChatGPT’s performance in order question.}
	\label{fig:chatgpt_performance}
\end{figure*}
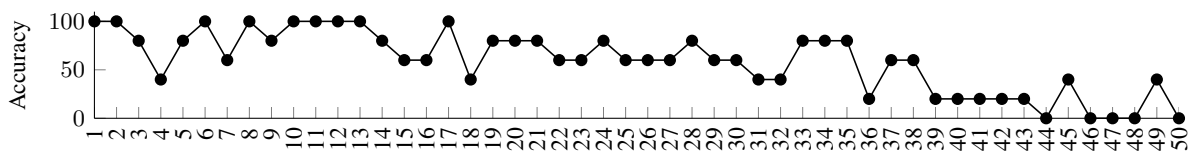

Figure~\ref{fig:chatgpt_performance} illustrates the average number of right responses given by ChatGPT for each question across all years. The data exhibits that the possibility of ChatGPT providing an accurate response reduces as the question's level of complexity rises. The ChatGPT correct answer rate is greater than 50\% for questions 1 through 35, which are K and C-level questions. The accurate answer rate of ChatGPT, however, decreases below 50\% for questions 35 to 50, demonstrating a decline proportional to the pattern of the questions. The graph demonstrates that as question difficulty grows, ChatGPT's accuracy declines. Given that questions at higher knowledge levels tend to be more complicated and need in-depth comprehension and problem-solving abilities, this pattern is to be expected. The findings imply that the difficulty and complexity of the questions have a significant impact on ChatGPT's capacity to provide accurate answers. This discovery has significant implications for the design of AI systems for educational applications since it emphasizes the need for more sophisticated and advanced models that are capable of handling difficult and challenging tasks. Additionally, it suggests that more investigation is required to identify the specific factors that influence ChatGPT's performance on various question types. This understanding can guide the creation of more efficient AI-based educational tools and interventions.
	
The analysis of the model's performance in relation to the order of the questions can be beneficial in a number of ways, in addition to determining ChatGPT's accuracy in responding to the questions. In the first place, it can assist teachers in comprehending how the order of questions impacts ChatGPT's capacity to solve them and in optimizing the question sequence to produce a more useful evaluation. This is crucial because as an exam goes on, students may become cognitively fatigued, which may affect how well they perform on subsequent questions. Teachers can simulate how students could perform under various circumstances and create exams that are better suited to accurately assess their knowledge and abilities by studying ChatGPT's performance with regard to the configuration of questions. Understanding how the question sequence impacts ChatGPT's performance can also assist identify possible weak points in the model, which can guide future model improvements.

\subsection{ChatGPT’s performance in levels and topics }

According to the degree of difficulty, Table~\ref{tabl:performance_in_levels} shows the percentage of accurate responses using ChatGPT for each year. The average percentage of right answers for K-level questions given by ChatGPT ranged from 90\% in 2022 to 75\% in 2023. The highest percentage of accurate answers for C-level questions was 75.22\% in 2022, and the lowest was 40\% in 2023. The highest and lowest percentages of right responses for questions at the A-level were 55.56\% and 0\%, respectively. For the years 2021, 2022, and 2023, ChatGPT did not offer any accurate responses to H-type questions. The highest percentages for the remaining years were 16.67\% and 22.22\%. These results show how ChatGPT has performed over time at various levels of difficulty.

\begingroup
\renewcommand{\arraystretch}{1.5} 
	\begin{table}[ht!]
		\centering
		\caption{ChatGPT’s performance in question levels }
		\label{tabl:performance_in_levels}
	\begin{tabular}{c|c|c|c|c|}
		\cline{2-5}
		\multicolumn{1}{l|}{}               & \textbf{K} & \textbf{C} & \textbf{A} & \textbf{H} \\ \hline
		\multicolumn{1}{|c|}{\textbf{2023}} & 75.00      & 40.00      & 25.00      & 0.00       \\ \hline
		\multicolumn{1}{|c|}{\textbf{2022}} & 90.00      & 72.22      & 0.00       & 0.00       \\ \hline
		\multicolumn{1}{|c|}{\textbf{2021}} & 81.82      & 62.50      & 28.57      & 0.00       \\ \hline
		\multicolumn{1}{|c|}{\textbf{2020}} & 89.47      & 62.50      & 55.56      & 16.67      \\ \hline
		\multicolumn{1}{|c|}{\textbf{2019}} & 85.71      & 58.82      & 20.00      & 22.22      \\ \hline
	\end{tabular}
\end{table}

\endgroup

\begin{figure*}[h]
	\begin{center}
		\begin{tikzpicture}
			\pgfplotstableread[col sep=comma]{level}\csvdata
			\pgfplotstabletranspose\datatransposed{\csvdata} 
			\begin{axis}[
				boxplot/draw direction = y,
				x axis line style = {opacity=0},
				axis x line* = bottom,
				height=6cm,
				axis y line = left,
				enlarge y limits,
				ymajorgrids,
				xtick = {1, 2, 3, 4},
				xticklabel style = {align=center, font=\small, rotate=0},
				xticklabels = {K, C, A, H},
				xtick style = {draw=none}, 
				ylabel = {Performance (\%)},
				ytick = {0, 25, 50, 75, 100}
				]
				\foreach \n in {1,...,4} {
					\addplot+[boxplot, fill, draw=black] table[y index=\n] {\datatransposed};
				}
			\end{axis}
		\end{tikzpicture}
	\end{center}
	\caption{ChatGPT’s performance in question levels for 2019-2023.}
	\label{fig:chatgpt_performance_lv}
\end{figure*}
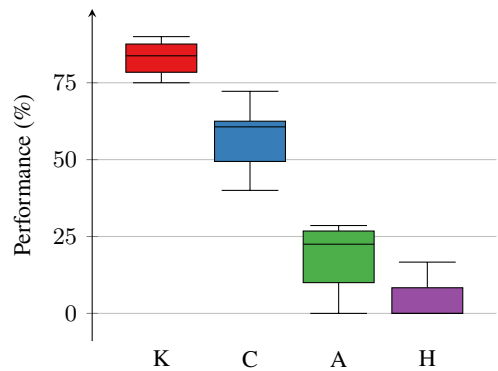

In accordance with the questions' degree of complexity, Figure~\ref{fig:chatgpt_performance_lv} depicts ChatGPT's accuracy from 2019 to 2023. For queries classified as type K, it indicates that ChatGPT attained an accuracy rate ranging from 75$\%$ to 90$\%$, with a small standard deviation indicating a high rate of consistency. This demonstrates ChatGPT's exceptional skill in answering questions that are not too challenging. For questions of type C, the accuracy rate falls to 40-72$\%$, demonstrating that ChatGPT performs less effectively when answering questions of intermediate difficulty. Type A questions show the greatest diversity in ChatGPT's accuracy rate, with correct answers ranging from 0$\%$ to 57$\%$ and the highest standard deviation. This shows that ChatGPT performs the least consistently when attempting to answer challenging type-A questions. The accuracy of ChatGPT's answers to the most difficult type H questions ranges from 0 to 22$\%$, which is a quite low percentage. Based on these findings, it appears that ChatGPT performs better when answering questions that are easier to answer than those that are more complex.

The percentage of correct responses offered by ChatGPT for different topics from 2019 to 2023 is depicted in Table~\ref{tabl:performance_topic}. ChatGPT provided 100\% accurate responses for all years for the topic M11B. Additionally, ChatGPT provided 100\% accurate responses for topics M11A, M12D, M12F, and M11C for a number of years. In 2022, ChatGPT's accuracy rate for the M11C topic was 0\%. With the exception of the M12A topic on graphs and diagrams, ChatGPT's accuracy rate for the other topics was rather high.

\begingroup
\renewcommand{\arraystretch}{1.5} 
\begin{center}
	\begin{table}[ht!]
		\caption{ChatGPT’s performance in question topics }
		\label{tabl:performance_topic}
		\resizebox{\textwidth}{!}{%

	\begin{tabular}{c|c|c|c|c|c|c|c|c|c|c|}
		\cline{2-11}
		\multicolumn{1}{l|}{}               & \textbf{M11C} & \textbf{M11B} & \textbf{M11A} & \textbf{M12A} & \textbf{M12B} & \textbf{M12C} & \textbf{M12D} & \textbf{M12E} & \textbf{M12F} & \textbf{M12G} \\ \hline
		\multicolumn{1}{|c|}{\textbf{2023}} & 50            & 100.00        & 50.00         & 30.00         & 75.00         & 57.14         & 83.33         & 33.33         & 50.00         & 44.44         \\ \hline
		\multicolumn{1}{|c|}{\textbf{2022}} & 0             & 100.00        & 50.00         & 50.00         & 75.00         & 71.43         & 66.67         & 66.67         & 66.67         & 62.50         \\ \hline
		\multicolumn{1}{|c|}{\textbf{2021}} & 50            & 100.00        & 100.00        & 20.00         & 75.00         & 71.43         & 66.67         & 66.67         & 66.67         & 62.50         \\ \hline
		\multicolumn{1}{|c|}{\textbf{2020}} & 100           & 100.00        & 100.00        & 46.15         & 62.50         & 42.86         & 100.00        & 66.67         & 100.00        & 75.00         \\ \hline
		\multicolumn{1}{|c|}{\textbf{2019}} &               & 100.00        & 50.00         & 28.57         & 71.43         & 80.00         & 40.00         & 80.00         & 33.33         & 50.00         \\ \hline
	\end{tabular}
	}
\end{table}
\end{center}
\endgroup

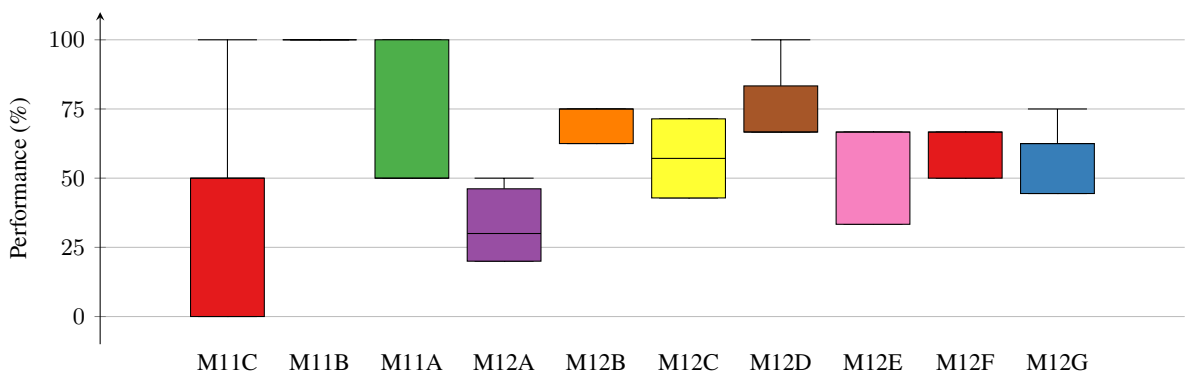
\begin{figure*}[h]
	\begin{center}
		\begin{tikzpicture}
			\pgfplotstableread[col sep=comma]{topic}\csvdata
			\pgfplotstabletranspose\datatransposed{\csvdata} 
			\begin{axis}[
				boxplot/draw direction = y,
				x axis line style = {opacity=0},
				axis x line* = bottom,
				height=6cm, width=16cm,
				axis y line = left,
				enlarge y limits,
				ymajorgrids,
				xtick = {1, 2, 3, 4, 5, 6, 7, 8, 9, 10},
				xticklabel style = {align=center, font=\small, rotate=0},
				xticklabels = {M11C, M11B, M11A, M12A, M12B, M12C, M12D,	M12E,	M12F,	M12G
				},
				xtick style = {draw=none}, 
				ylabel = {Performance (\%)},
				ytick = {0, 25, 50, 75, 100}
				]
				\foreach \n in {1,...,10} {
					\addplot+[boxplot, fill, draw=black] table[y index=\n] {\datatransposed};
				}
			\end{axis}
		\end{tikzpicture}
	\end{center}
	\caption{ChatGPT’s performance in question topics for 2019-2023.}
	\label{fig:chatgpt_performance_topics}
\end{figure*}

Recently, a lot of attention has been paid to how well AI models perform, particularly when answering questions. Figure~\ref{fig:chatgpt_performance_topics} provides an informative examination of ChatGPT's accuracy in responding to various query kinds over the period of 2019–2023. The findings show that ChatGPT's accuracy varies depending on the type of question being answered. In particular, ChatGPT answered M11C questions with an accuracy rate of 0–100$\%$, M11B questions with 100$\%$, M11A questions with 50–100$\%$, M12A questions with 20–50$\%$, M12B questions with 62–75$\%$, M12C questions with 42–80$\%$, M12D questions with 40–100$\%$, M12E questions with 33–80$\%$, M12F questions with 33–100$\%$, and M12G questions with 44–75$\%$.

The level of difficulty of the questions, the number and quality of training data, and the model's internal architecture are just a few of the variables that can affect how well ChatGPT performs while answering these questions. Therefore, comprehending the variations in performance across various question types can offer insights into the model's advantages and disadvantages as well as guide future developments to enhance its performance.

A thorough analysis of ChatGPT's performance on various levels and topics is presented in Table~\ref{tabl:ChatGPT’s performance in knowledge matrix}. First, consider the difficulty of the questions; ChatGPT was able to accurately respond to 85 of 103 questions at level K. Out of 77 questions at level C, 48 were correctly answered by ChatGPT. Only 12 of the 49 questions in level A could be correctly answered by ChatGPT, while only 3 of the 29 questions in level H could be answered by ChatGPT. Second, ChatGPT's performance varied depending on the type of question. For M11A, M11B, M11C, and M12A, ChatGPT correctly answered 7 out of 10 questions, 5 out of 5 questions, 4 out of 8 questions, and 20 out of 57 questions, respectively. For M12B, M12C, M12D, M12E, M12F, and M12G, respectively, ChatGPT correctly answered 28 out of 39 questions, 21 out of 33 questions, 18 out of 26 questions, 11 out of 16 questions, 9 out of 15 questions, and 24 out of 41 questions. 

It is crucial to keep in mind that certain topics only contain questions at the knowledge and comprehension levels that are quite simple to respond to, and ChatGPT did well on these because of its aptitude for natural language creation. Therefore, ChatGPT's high scores on these topics do not necessarily reflect its understanding of mathematics or capacity for reasoning. Furthermore, it is challenging to give a precise rating solely based on topics because some topics have a preponderance of knowledge-level questions. Additionally, due to a lack of information, ChatGPT might not be able to respond to some knowledge-level questions. As an illustration, many questions in the topic of derivatives and applications (M12A) call for the interpretation of graphs or variable tables, which ChatGPT is unable to read from photos at this time. As a result, ChatGPT might be unable to respond to some inquiries that require an understanding of this subject. These findings show that ChatGPT has diverse degrees of competence in various math specialties. In general, ChatGPT performed well for some question types but poorly for others.

\begingroup
\renewcommand{\arraystretch}{1.5} 
\begin{center}
	\begin{table}[ht!]
		\caption{ChatGPT’s performance in knowledge matrix}
		\label{tabl:ChatGPT’s performance in knowledge matrix}
		\resizebox{\textwidth}{!}{%

	\begin{tabular}{c|c|c|c|c|c|c|c|c|c|c|cc}
		\cline{2-11}
		& \textbf{M11C} & \textbf{M11B} & \textbf{M11A} & \textbf{M12A} & \textbf{M12B} & \textbf{M12C} & \textbf{M12D} & \textbf{M12E} & \textbf{M12F} & \textbf{M12G} & \multicolumn{2}{c}{\textbf{LEVEL}}                                       \\ \hline
		\multicolumn{1}{|c|}{\textbf{K}} & 1             & 5             & 5             & 12            & 15            & 12            & 8             & 7             & 7             & 13            & \multicolumn{1}{c|}{85}           & \multicolumn{1}{c|}{83$\%$}             \\ \hline
		\multicolumn{1}{|c|}{\textbf{C}} & 2             &               & 1             & 6             & 11            & 7             & 8             & 2             & 1             & 10            & \multicolumn{1}{c|}{48}           & \multicolumn{1}{c|}{62$\%$}             \\ \hline
		\multicolumn{1}{|c|}{\textbf{A}} & 1             &               & 1             & 0             & 2             & 2             & 2             & 1             & 1             & 1             & \multicolumn{1}{c|}{11}           & \multicolumn{1}{c|}{27$\%$}             \\ \hline
		\multicolumn{1}{|c|}{\textbf{H}} & 0             &               &               & 2             & 0             & 0             & 0             & 1             &               & 0             & \multicolumn{1}{c|}{3}            & \multicolumn{1}{c|}{10$\%$}             \\ \hline
		\multirow{2}{*}{\textbf{TOPIC}}  & 4             & 5             & 7             & 20            & 28            & 21            & 18            & 11            & 9             & 24            & \multicolumn{1}{c|}{\textbf{147}} &                                      \\ \cline{2-13} 
		& 50$\%$           & 100$\%$          & 70$\%$           & 35$\%$           & 72$\%$           & 64$\%$           & 69$\%$           & 65$\%$           & 64$\%$           & 59$\%$           & \multicolumn{1}{c|}{}             & \multicolumn{1}{c|}{\textbf{58.80$\%$}} \\ \cline{2-11} \cline{13-13} 
	\end{tabular}

	}
\end{table}
\end{center}
\endgroup

\begin{figure*}[h]
	\begin{center}
		\includegraphics[width=\textwidth]{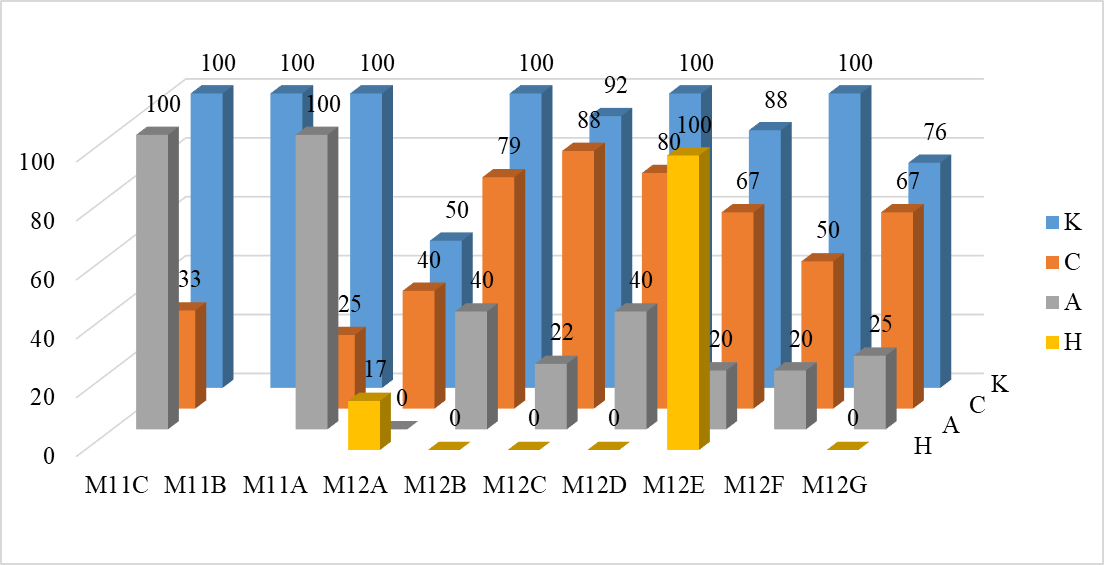}
	\end{center}
	\caption{Distribution of the percentage of correct answer in levels and topics.}
	\label{fig:Distribution of the percentage of correct answer in levels and topics}
\end{figure*}

These results collectively imply that while ChatGPT might be a valuable tool for addressing math-related queries, its accuracy varies between topics and levels. As a result, significant advancements are required to increase ChatGPT's math question-answering ability, especially in more difficult math subfields. Figure~\ref{fig:Distribution of the percentage of correct answer in levels and topics} presents a more thorough breakdown of the percentage of right responses by difficulty level and topic so that users of ChatGPT can better understand how well it performs. For instance, in the case of M12G, ChatGPT attained a high accuracy rate of 76$\%$ for questions at the K level, followed by 67$\%$ for questions at the C level, 25$\%$ for questions at the A level, and 0$\%$ for questions at the H level. Notably, ChatGPT achieved a flawless accuracy rate of 100$\%$ when responding to questions at the K level for M11A, M11B, M11C, M12B, M12D, and M12F. Additionally, ChatGPT was able to correctly respond to H-level questions for M12A (Derivatives and Applications) and M12E (Polyhedron), demonstrating its competency in handling more difficult questions in these topics. These results indicate that the topic and difficulty level have an impact on ChatGPT's accuracy, and that ChatGPT performs differently depending on how these two factors are coupled.  These findings suggest that these particular issues contain linguistic nuances or complexities that the model was unable to adequately capture. This result highlights the need for ongoing study to enhance the model's ability to handle a variety of linguistic complexities. This shortcoming might be brought on by the lack of training data or the intrinsic intricacy of the queries at this level.

By evaluating how well language models—like ChatGPT—can respond to questions of varying degrees of cognitive complexity, one can assess the performance of these models. Knowledge, understanding, application, and strong application are the four categories for the levels of cognitive difficulty in answering questions. The ability to recognize and identify concepts, content, and issues is referred to as the recognition level. Understanding fundamental ideas and being able to articulate them in one's own words are requirements for the comprehension level. The application level necessitates applying concepts in unfamiliar or comparable circumstances. The high application level requires the capacity to apply fundamental ideas to an entirely new challenge.

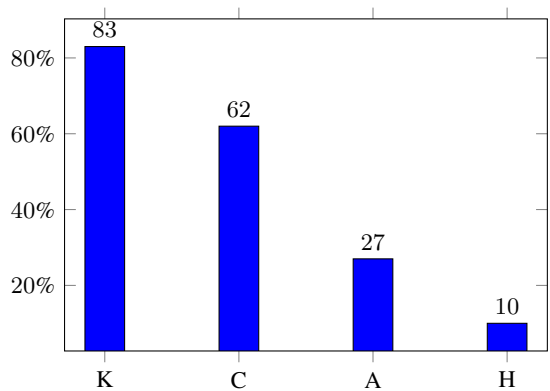
\begin{figure*}[h]
	\begin{center}
		\begin{tikzpicture}  
			\begin{axis}  
				[  
				ybar, 
				bar width=15pt, 
				legend style={at={(0.5,-0.275)}, 	
					anchor=north,legend columns=-1},    
				symbolic x coords={K, C, A, H}, 
				xtick=data,
				yticklabel={\pgfmathprintnumber{\tick}\%},  
				nodes near coords,   
				nodes near coords align={vertical}, 
				enlarge x limits,
				height=6cm, width=8cm, 
				]  
				\addplot [fill=blue] coordinates {
					(K, 83)
					(C, 62)
					(A, 27)
					(H, 10)
				}; 
			\end{axis}  
		\end{tikzpicture}
	\end{center}
	\caption{ChatGPT’s performance in question levels.}
	\label{fig:ChatGPT’s performance in question level}
\end{figure*}

The effectiveness of ChatGPT was assessed by counting how many questions at each level of cognitive difficulty it correctly answered. Figure~\ref{fig:ChatGPT’s performance in question level} demonstrates that ChatGPT properly identified and recognized 83$\%$ of the ideas in the recognition level of the questions that were asked. 62$\%$ of the questions at the comprehension level were correctly answered by ChatGPT, demonstrating an adequate understanding of the fundamental ideas. At the application level, where it could only accurately answer 27$\%$ of the questions, its performance deteriorated dramatically. Only 10$\%$ of the questions were correctly answered by ChatGPT at the highest cognitive complexity level, the high application level, demonstrating a limited capacity to apply fundamental ideas to novel problems.

According to this performance evaluation, ChatGPT may have some restrictions when it comes to employing newly learned concepts in novel contexts. By giving language models more sophisticated and advanced problem-solving abilities, future language model development might concentrate on enhancing the models' capacity to solve novel challenges. The performance of language models at the application and high application levels may also be enhanced by additional training data and focused training techniques, enabling them to more effectively apply acquired concepts in real-world circumstances.

\begin{figure*}[h]
	\begin{center}
		\begin{tikzpicture}
			\begin{axis}[
				symbolic y coords={
					M12A,
					M11C,
					M12G,
					M12F,
					M12C,
					M12E,
					M11A,
					M12D,
					M12B,
					M11B,
				},
				ytick=data,
				xbar,
				bar width=5pt,
				xmin=0,
				enlarge y limits,
				height=6cm, 
				width=0.5\textwidth,
				axis x line*=bottom, axis y line*=left
				]
				
				\addplot coordinates {
					(34.9450549450549,M12A)
					(50,M11C)
					(58.8888888888889,M12G)
					(60,M12F)
					(64.5714285714286,M12C)
					(66,M12E)
					(70,M11A)
					(71.3333333333333,M12D)
					(71.7857142857143,M12B)
					(100,M11B)
				};
			\end{axis}
		\end{tikzpicture}
	\end{center}
	\caption{ChatGPT’s performance in question topics.}
	\label{fig:. ChatGPT’s performance in question type}
\end{figure*}
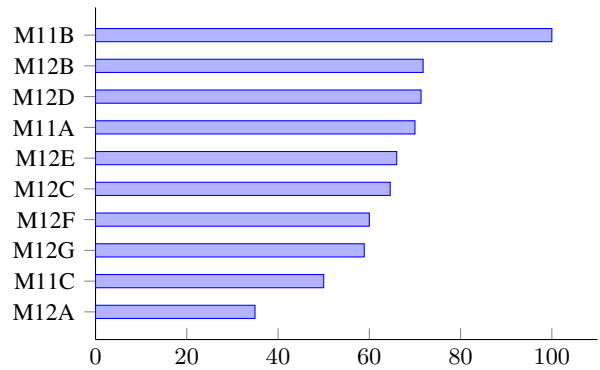

Figure~\ref{fig:. ChatGPT’s performance in question type} demonstrates the astounding 100$\%$ correct answer rate for the M11B question that ChatGPT attained. It's crucial to remember that this particular topic only included K-type questions. The correct answer rates for the remaining topics ranged from 58.89$\%$ for M12G to 71.79$\%$ for M12B. Notably, M11C and M12A had the lowest rates of correctly answered questions. Most questions were in M12A, and the majority of them were at the K-level. The lack of information in the figure, however, prevented ChatGPT from being able to respond to all questions. Similarly, ChatGPT did not show much promise for topics like M11C on spatial geometry and M12G on spatial analysis Oxyz. 

However, if we ignore the questions that required information from the figure, ChatGPT demonstrated a solid capacity to respond correctly for more than 50$\%$ of all topics. This indicates that ChatGPT shows potential in some areas of the evaluated topics, but it may need more work to succeed in other areas that require more intricate inference and data interpretation.

\subsection{ChatGPT’s performance in VNHSGE and other exams }

\begin{figure*}[h]
	\centering
		\begin{tikzpicture}
		\begin{axis}[
			symbolic y coords={
				AP Calculus BC,
				AMC 12,
				AMC 10,
				GRE Quantitative,
				AP Statistics,
				VNHSGE Mathematics,
				SAT Math
			},
			ytick=data,
			xbar,
			bar width=5pt,
			xmin=0,
			enlarge y limits,
			height=5cm, 
			width=0.5\textwidth,
			axis x line*=bottom, axis y line*=left
			]
			
			\addplot coordinates {
				(70,SAT Math)
				(58.8,VNHSGE Mathematics)
				(40,AP Statistics)
				(25,GRE Quantitative)
				(10,AMC 10)
				(4,AMC 12)
				(1,AP Calculus BC)
			};
		\end{axis}
	\end{tikzpicture}
	\caption{ChatGPT’s performance in VNHSGE mathematics and other exams.}
	\label{fig:ChatGPT’s performance in VNHSGE Math and other exams}
\end{figure*}
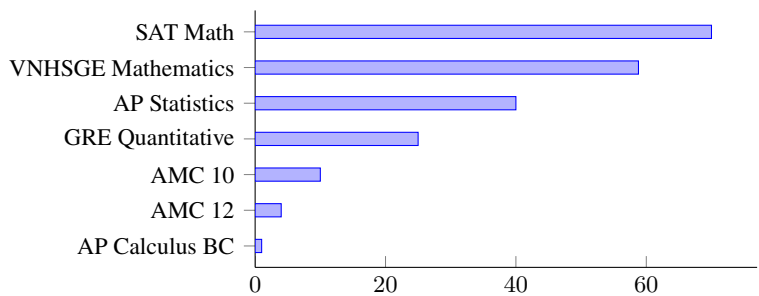

We evaluated ChatGPT's success rate in a number of well-known math competitions, as reported by OpenAI~\cite{OpenAI_gpt_4_report} and shown in Figure~\ref{fig:ChatGPT’s performance in VNHSGE Math and other exams}, to determine its suitability for the VNHSGE mathematics exam. With a success percentage of 70$\%$, ChatGPT's performance in the SAT Math competition is better than its performance in the VNHSGE mathematics exam, according to our study. With rates of 40$\%$ for AP Statistics, 25$\%$ for the GRE Quantitative, 10$\%$ for AMC 10, 4$\%$ for AMC 12, and only 1$\%$ for AP Calculus BC, ChatGPT performed much worse in the other competitions. It is important to note that these comparisons are just meant to be used as a guide because there are variations among math examinations in terms of their formats, structures, levels, and question kinds. As a result, it is impossible to assess the complexity of the VNHSGE exam just by looking at ChatGPT's performance in other competitions. However, this comparison provides a general idea of the VNHSGE exam's level of difficulty in relation to other math competitions.

\subsection{ChatGPT’s performance and Vietnamese students  }

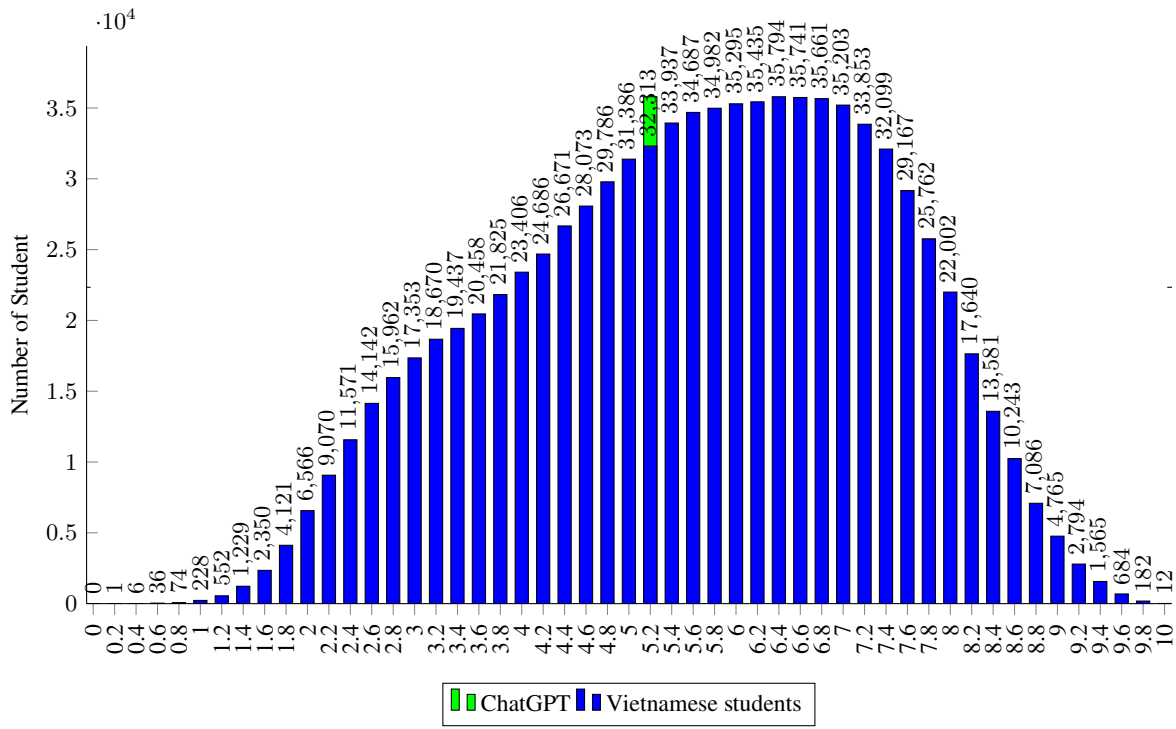
\begin{figure}[h!]
	\begin{center}
		\begin{tikzpicture}
			\begin{axis}[
				legend style={at={(0.5,-0.125)}, 	
					anchor=north,legend columns=-1}, 
				symbolic x coords={
					0,
					0.2,
					0.4,
					0.6,
					0.8,
					1,
					1.2,
					1.4,
					1.6,
					1.8,
					2,
					2.2,
					2.4,
					2.6,
					2.8,
					3,
					3.2,
					3.4,
					3.6,
					3.8,
					4,
					4.2,
					4.4,
					4.6,
					4.8,
					5,
					5.2,
					5.4,
					5.6,
					5.8,
					6,
					6.2,
					6.4,
					6.6,
					6.8,
					7,
					7.2,
					7.4,
					7.6,
					7.8,
					8,
					8.2,
					8.4,
					8.6,
					8.8,
					9,
					9.2,
					9.4,
					9.6,
					9.8,
					10,
				},
				hide axis,
				ybar,
				bar width=5pt,
				ymin=0,
				every node near coord/.append style={rotate=90, anchor=west},
				width=\textwidth, 
				enlarge x limits={abs=0.5*\pgfplotbarwidth},
				height=10cm, 
				width=16cm,
				axis x line*=bottom, axis y line*=left
				]
				\addplot [fill=green] coordinates {
					(0,0)
				};
				\addplot [fill=blue] coordinates {
					(10,0)
				};	
				\legend{ChatGPT, Vietnamese students }	
			\end{axis}
			
			\begin{axis}[
				symbolic x coords={
					0,
					0.2,
					0.4,
					0.6,
					0.8,
					1,
					1.2,
					1.4,
					1.6,
					1.8,
					2,
					2.2,
					2.4,
					2.6,
					2.8,
					3,
					3.2,
					3.4,
					3.6,
					3.8,
					4,
					4.2,
					4.4,
					4.6,
					4.8,
					5,
					5.2,
					5.4,
					5.6,
					5.8,
					6,
					6.2,
					6.4,
					6.6,
					6.8,
					7,
					7.2,
					7.4,
					7.6,
					7.8,
					8,
					8.2,
					8.4,
					8.6,
					8.8,
					9,
					9.2,
					9.4,
					9.6,
					9.8,
					10,
				},
				hide axis,
				x tick label style={rotate=90,anchor=east},
				ybar,
				bar width=5pt,
				ymin=0,
				every node near coord/.append style={rotate=90, anchor=west},
				width=\textwidth, 
				enlarge x limits={abs=0.5*\pgfplotbarwidth},
				height=9cm, 
				width=16cm,
				axis x line*=bottom, axis y line*=left
				]
				\addplot [fill=green] coordinates {
					(0,0)
					(0.2,0)
					(0.4,0)
					(0.6,0)
					(0.8,0)
					(1,0)
					(1.2,0)
					(1.4,0)
					(1.6,0)
					(1.8,0)
					(2,0)
					(2.2,0)
					(2.4,0)
					(2.6,0)
					(2.8,0)
					(3,0)
					(3.2,0)
					(3.4,0)
					(3.6,0)
					(3.8,0)
					(4,0)
					(4.2,0)
					(4.4,0)
					(4.6,0)
					(4.8,0)
					(5,0)
					(5.2,55000)
					(5.4,0)
					(5.6,0)
					(5.8,0)
					(6,0)
					(6.2,0)
					(6.4,0)
					(6.6,0)
					(6.8,0)
					(7,0)
					(7.2,0)
					(7.4,0)
					(7.6,0)
					(7.8,0)
					(8,0)
					(8.2,0)
					(8.4,0)
					(8.6,0)
					(8.8,0)
					(9,0)
					(9.2,0)
					(9.4,0)
					(9.6,0)
					(9.8,0)
					(10,0)
					
				};	
			\end{axis}
			
			\begin{axis}[
				ylabel={Number of Student},
				symbolic x coords={
					0,
					0.2,
					0.4,
					0.6,
					0.8,
					1,
					1.2,
					1.4,
					1.6,
					1.8,
					2,
					2.2,
					2.4,
					2.6,
					2.8,
					3,
					3.2,
					3.4,
					3.6,
					3.8,
					4,
					4.2,
					4.4,
					4.6,
					4.8,
					5,
					5.2,
					5.4,
					5.6,
					5.8,
					6,
					6.2,
					6.4,
					6.6,
					6.8,
					7,
					7.2,
					7.4,
					7.6,
					7.8,
					8,
					8.2,
					8.4,
					8.6,
					8.8,
					9,
					9.2,
					9.4,
					9.6,
					9.8,
					10,
				},
				xtick=data,
				x tick label style={rotate=90,anchor=east},
				ybar,
				bar width=5pt,
				ymin=0,
				nodes near coords,   
				every node near coord/.append style={rotate=90, anchor=west},
				width=\textwidth, 
				enlarge x limits={abs=0.5*\pgfplotbarwidth},
				height=9cm, 
				width=16cm,
				axis x line*=bottom, axis y line*=left
				]
				\addplot [fill=blue] coordinates {
					(0,0)
					(0.2,1)
					(0.4,6)
					(0.6,36)
					(0.8,74)
					(1,228)
					(1.2,552)
					(1.4,1229)
					(1.6,2350)
					(1.8,4121)
					(2,6566)
					(2.2,9070)
					(2.4,11571)
					(2.6,14142)
					(2.8,15962)
					(3,17353)
					(3.2,18670)
					(3.4,19437)
					(3.6,20458)
					(3.8,21825)
					(4,23406)
					(4.2,24686)
					(4.4,26671)
					(4.6,28073)
					(4.8,29786)
					(5,31386)
					(5.2,32313)
					(5.4,33937)
					(5.6,34687)
					(5.8,34982)
					(6,35295)
					(6.2,35435)
					(6.4,35794)
					(6.6,35741)
					(6.8,35661)
					(7,35203)
					(7.2,33853)
					(7.4,32099)
					(7.6,29167)
					(7.8,25762)
					(8,22002)
					(8.2,17640)
					(8.4,13581)
					(8.6,10243)
					(8.8,7086)
					(9,4765)
					(9.2,2794)
					(9.4,1565)
					(9.6,684)
					(9.8,182)
					(10,12)
				};	
				
			\end{axis}
		\end{tikzpicture}
	\end{center}
	\caption{Mathematics score spectrum of Vietnamese students in 2019.}
	\label{fig:math_2019}
\end{figure}

Figure~\ref{fig:math_2019}-\ref{fig:math_2023} compare ChatGPT math scores across four years—specifically, 2019, 2020, 2021, 2022 and 2023—with Vietnamese students' scores. Notably, the findings show that across the investigated years, ChatGPT math scores have consistently been lower than those of the majority of Vietnamese pupils. Additional performance data analysis can shed light on potential causes of the performance gap between ChatGPT and human students. There may be a variance in performance due to elements such various learning styles and approaches, resource accessibility, and cultural background. Additionally, with additional training and model improvement, ChatGPT's performance might be enhanced.

Another key drawback of this AI model is ChatGPT's inability to access, read, and comprehend graphical information in test questions. Tables, charts, and other graphical representations of data and information are frequently used in mathematics exams to visually communicate data and information. However, ChatGPT's inability to interpret graphical data limits its capacity to offer precise answers to this kind of query.

This restriction is not specific to ChatGPT; many other AI models also have trouble comprehending graphical data. This is so because reading text takes a distinct set of abilities than analyzing images and other visual information. NLP is exploited by text-based AI models like ChatGPT to comprehend and process text-based inputs. In contrast, computer vision techniques are utilized by image-based AI models to comprehend visual inputs.

Enhancing ChatGPT's capacity to comprehend visual data is one potential means of getting around this restriction. Adding computer vision capabilities to the model or creating a hybrid model that blends NLP and computer vision methods may achieve this. The test format could be changed to eliminate graphical data or to offer alternate text-based representations of the graphical data as a potential alternative. Though it might not always be possible, this solution would necessitate significant modifications to the test design.

\begin{figure}[h!]
	\begin{center}
		\begin{tikzpicture}
			\begin{axis}[
				legend style={at={(0.5,-0.125)}, 	
					anchor=north,legend columns=-1}, 
				symbolic x coords={
					0,
					0.2,
					0.4,
					0.6,
					0.8,
					1,
					1.2,
					1.4,
					1.6,
					1.8,
					2,
					2.2,
					2.4,
					2.6,
					2.8,
					3,
					3.2,
					3.4,
					3.6,
					3.8,
					4,
					4.2,
					4.4,
					4.6,
					4.8,
					5,
					5.2,
					5.4,
					5.6,
					5.8,
					6,
					6.2,
					6.4,
					6.6,
					6.8,
					7,
					7.2,
					7.4,
					7.6,
					7.8,
					8,
					8.2,
					8.4,
					8.6,
					8.8,
					9,
					9.2,
					9.4,
					9.6,
					9.8,
					10,
				},
				hide axis,
				ybar,
				bar width=5pt,
				ymin=0,
				every node near coord/.append style={rotate=90, anchor=west},
				width=\textwidth, 
				enlarge x limits={abs=0.5*\pgfplotbarwidth},
				height=10cm, 
				width=16cm,
				axis x line*=bottom, axis y line*=left
				]
				\addplot [fill=green] coordinates {
					(0,0)
				};
				\addplot [fill=blue] coordinates {
					(10,0)
				};	
				\legend{ChatGPT, Vietnamese students }	
			\end{axis}
			
			\begin{axis}[
				symbolic x coords={
					0,
					0.2,
					0.4,
					0.6,
					0.8,
					1,
					1.2,
					1.4,
					1.6,
					1.8,
					2,
					2.2,
					2.4,
					2.6,
					2.8,
					3,
					3.2,
					3.4,
					3.6,
					3.8,
					4,
					4.2,
					4.4,
					4.6,
					4.8,
					5,
					5.2,
					5.4,
					5.6,
					5.8,
					6,
					6.2,
					6.4,
					6.6,
					6.8,
					7,
					7.2,
					7.4,
					7.6,
					7.8,
					8,
					8.2,
					8.4,
					8.6,
					8.8,
					9,
					9.2,
					9.4,
					9.6,
					9.8,
					10,
				},
				hide axis,
				x tick label style={rotate=90,anchor=east},
				ybar,
				bar width=5pt,
				ymin=0,
				every node near coord/.append style={rotate=90, anchor=west},
				width=\textwidth, 
				enlarge x limits={abs=0.5*\pgfplotbarwidth},
				height=9cm, 
				width=16cm,
				axis x line*=bottom, axis y line*=left
				]
				\addplot [fill=green] coordinates {
					(0,0)
					(0.2,0)
					(0.4,0)
					(0.6,0)
					(0.8,0)
					(1,0)
					(1.2,0)
					(1.4,0)
					(1.6,0)
					(1.8,0)
					(2,0)
					(2.2,0)
					(2.4,0)
					(2.6,0)
					(2.8,0)
					(3,0)
					(3.2,0)
					(3.4,0)
					(3.6,0)
					(3.8,0)
					(4,0)
					(4.2,0)
					(4.4,0)
					(4.6,0)
					(4.8,0)
					(5,0)
					(5.2,0)
					(5.4,0)
					(5.6,0)
					(5.8,0)
					(6,0)
					(6.2,0)
					(6.4,0)
					(6.6,55000)
					(6.8,0)
					(7,0)
					(7.2,0)
					(7.4,0)
					(7.6,0)
					(7.8,0)
					(8,0)
					(8.2,0)
					(8.4,0)
					(8.6,0)
					(8.8,0)
					(9,0)
					(9.2,0)
					(9.4,0)
					(9.6,0)
					(9.8,0)
					(10,0)
					
				};	
			\end{axis}
			\begin{axis}[
				ylabel={Number of Student},
				symbolic x coords={
					0,
					0.2,
					0.4,
					0.6,
					0.8,
					1,
					1.2,
					1.4,
					1.6,
					1.8,
					2,
					2.2,
					2.4,
					2.6,
					2.8,
					3,
					3.2,
					3.4,
					3.6,
					3.8,
					4,
					4.2,
					4.4,
					4.6,
					4.8,
					5,
					5.2,
					5.4,
					5.6,
					5.8,
					6,
					6.2,
					6.4,
					6.6,
					6.8,
					7,
					7.2,
					7.4,
					7.6,
					7.8,
					8,
					8.2,
					8.4,
					8.6,
					8.8,
					9,
					9.2,
					9.4,
					9.6,
					9.8,
					10,
				},
				xtick=data,
				x tick label style={rotate=90,anchor=east},
				ybar,
				bar width=5pt,
				ymin=0,
				nodes near coords,   
				every node near coord/.append style={rotate=90, anchor=west},
				width=\textwidth, 
				enlarge x limits={abs=0.5*\pgfplotbarwidth},
				height=9cm, 
				width=16cm,
				axis x line*=bottom, axis y line*=left
				]
				\addplot [fill=blue] coordinates {
					(0,1)
					(0.2,1)
					(0.4,3)
					(0.6,9)
					(0.8,47)
					(1,134)
					(1.2,292)
					(1.4,668)
					(1.6,1212)
					(1.8,2189)
					(2,3092)
					(2.2,4421)
					(2.4,5642)
					(2.6,6792)
					(2.8,7725)
					(3,8452)
					(3.2,9190)
					(3.4,9645)
					(3.6,10573)
					(3.8,11330)
					(4,12248)
					(4.2,13107)
					(4.4,14400)
					(4.6,15475)
					(4.8,16719)
					(5,18136)
					(5.2,19609)
					(5.4,20981)
					(5.6,22363)
					(5.8,23502)
					(6,24943)
					(6.2,26287)
					(6.4,28088)
					(6.6,29894)
					(6.8,31927)
					(7,34273)
					(7.2,36783)
					(7.4,39596)
					(7.6,41075)
					(7.8,41868)
					(8,41567)
					(8.2,40297)
					(8.4,38260)
					(8.6,35625)
					(8.8,32792)
					(9,27237)
					(9.2,19433)
					(9.4,11086)
					(9.6,4669)
					(9.8,1542)
					(10,273)
					
				};	
				
			\end{axis}
		\end{tikzpicture}
	\end{center}
	\caption{Mathematics score spectrum of Vietnamese students in 2020.}
	\label{fig:math_2020}
\end{figure}
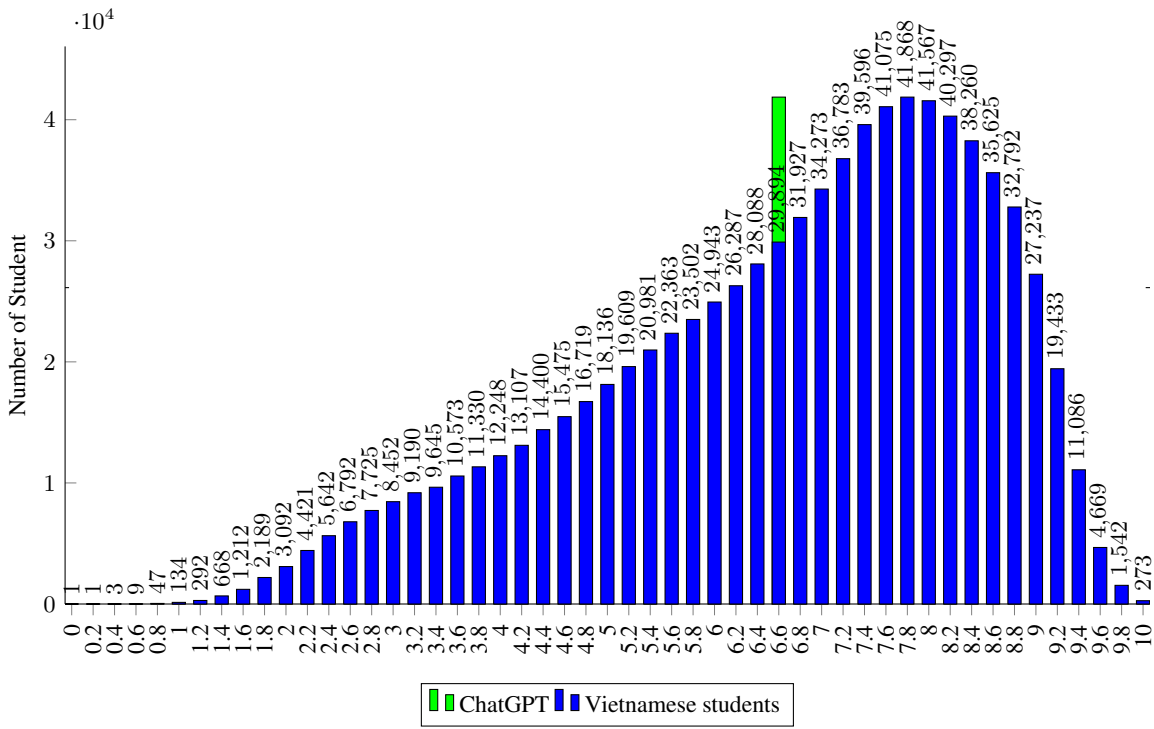

\begin{figure}[h!]
	\begin{center}
		\begin{tikzpicture}
			\begin{axis}[
				legend style={at={(0.5,-0.125)}, 	
					anchor=north,legend columns=-1}, 
				symbolic x coords={
					0,
					0.2,
					0.4,
					0.6,
					0.8,
					1,
					1.2,
					1.4,
					1.6,
					1.8,
					2,
					2.2,
					2.4,
					2.6,
					2.8,
					3,
					3.2,
					3.4,
					3.6,
					3.8,
					4,
					4.2,
					4.4,
					4.6,
					4.8,
					5,
					5.2,
					5.4,
					5.6,
					5.8,
					6,
					6.2,
					6.4,
					6.6,
					6.8,
					7,
					7.2,
					7.4,
					7.6,
					7.8,
					8,
					8.2,
					8.4,
					8.6,
					8.8,
					9,
					9.2,
					9.4,
					9.6,
					9.8,
					10,
				},
				hide axis,
				ybar,
				bar width=5pt,
				ymin=0,
				every node near coord/.append style={rotate=90, anchor=west},
				width=\textwidth, 
				enlarge x limits={abs=0.5*\pgfplotbarwidth},
				height=10cm, 
				width=16cm,
				axis x line*=bottom, axis y line*=left
				]
				\addplot [fill=green] coordinates {
					(0,0)
				};
				\addplot [fill=blue] coordinates {
					(10,0)
				};	
				\legend{ChatGPT, Vietnamese students }	
			\end{axis}
			
			\begin{axis}[
				symbolic x coords={
					0,
					0.2,
					0.4,
					0.6,
					0.8,
					1,
					1.2,
					1.4,
					1.6,
					1.8,
					2,
					2.2,
					2.4,
					2.6,
					2.8,
					3,
					3.2,
					3.4,
					3.6,
					3.8,
					4,
					4.2,
					4.4,
					4.6,
					4.8,
					5,
					5.2,
					5.4,
					5.6,
					5.8,
					6,
					6.2,
					6.4,
					6.6,
					6.8,
					7,
					7.2,
					7.4,
					7.6,
					7.8,
					8,
					8.2,
					8.4,
					8.6,
					8.8,
					9,
					9.2,
					9.4,
					9.6,
					9.8,
					10,
				},
				hide axis,
				x tick label style={rotate=90,anchor=east},
				ybar,
				bar width=5pt,
				ymin=0,
				every node near coord/.append style={rotate=90, anchor=west},
				width=\textwidth, 
				enlarge x limits={abs=0.5*\pgfplotbarwidth},
				height=9cm, 
				width=16cm,
				axis x line*=bottom, axis y line*=left
				]
				\addplot [fill=green] coordinates {
					(0,0)
					(0.2,0)
					(0.4,0)
					(0.6,0)
					(0.8,0)
					(1,0)
					(1.2,0)
					(1.4,0)
					(1.6,0)
					(1.8,0)
					(2,0)
					(2.2,0)
					(2.4,0)
					(2.6,0)
					(2.8,0)
					(3,0)
					(3.2,0)
					(3.4,0)
					(3.6,0)
					(3.8,0)
					(4,0)
					(4.2,0)
					(4.4,0)
					(4.6,0)
					(4.8,0)
					(5,0)
					(5.2,0)
					(5.4,0)
					(5.6,0)
					(5.8,0)
					(6,55000)
					(6.2,0)
					(6.4,0)
					(6.6,0)
					(6.8,0)
					(7,0)
					(7.2,0)
					(7.4,0)
					(7.6,0)
					(7.8,0)
					(8,0)
					(8.2,0)
					(8.4,0)
					(8.6,0)
					(8.8,0)
					(9,0)
					(9.2,0)
					(9.4,0)
					(9.6,0)
					(9.8,0)
					(10,0)
					
				};	
			\end{axis}
				
			\begin{axis}[
				ylabel={Number of Student},
				symbolic x coords={
					0,
					0.2,
					0.4,
					0.6,
					0.8,
					1,
					1.2,
					1.4,
					1.6,
					1.8,
					2,
					2.2,
					2.4,
					2.6,
					2.8,
					3,
					3.2,
					3.4,
					3.6,
					3.8,
					4,
					4.2,
					4.4,
					4.6,
					4.8,
					5,
					5.2,
					5.4,
					5.6,
					5.8,
					6,
					6.2,
					6.4,
					6.6,
					6.8,
					7,
					7.2,
					7.4,
					7.6,
					7.8,
					8,
					8.2,
					8.4,
					8.6,
					8.8,
					9,
					9.2,
					9.4,
					9.6,
					9.8,
					10,
				},
				xtick=data,
				x tick label style={rotate=90,anchor=east},
				ybar,
				bar width=5pt,
				ymin=0,
				nodes near coords,   
				every node near coord/.append style={rotate=90, anchor=west},
				width=\textwidth, 
				enlarge x limits={abs=0.5*\pgfplotbarwidth},
				height=9cm, 
				width=16cm,
				axis x line*=bottom, axis y line*=left
				]
				\addplot [fill=blue] coordinates {
					(0,1)
					(0.2,0)
					(0.4,0)
					(0.6,11)
					(0.8,22)
					(1,85)
					(1.2,199)
					(1.4,464)
					(1.6,856)
					(1.8,1488)
					(2,2370)
					(2.2,3379)
					(2.4,4613)
					(2.6,5929)
					(2.8,6920)
					(3,8145)
					(3.2,9450)
					(3.4,10673)
					(3.6,11987)
					(3.8,13454)
					(4,14986)
					(4.2,16519)
					(4.4,17928)
					(4.6,19593)
					(4.8,21730)
					(5,23301)
					(5.2,24943)
					(5.4,26711)
					(5.6,28011)
					(5.8,29725)
					(6,31210)
					(6.2,32877)
					(6.4,34974)
					(6.6,37229)
					(6.8,39978)
					(7,43491)
					(7.2,46401)
					(7.4,50532)
					(7.6,52947)
					(7.8,53972)
					(8,53133)
					(8.2,49929)
					(8.4,44855)
					(8.6,37943)
					(8.8,29562)
					(9,20147)
					(9.2,11097)
					(9.4,5049)
					(9.6,1647)
					(9.8,358)
					(10,52)		
				};	
				
			\end{axis}
		\end{tikzpicture}
	\end{center}
	\caption{Mathematics score spectrum of Vietnamese students in 2021.}
	\label{fig:math_2021}
\end{figure}
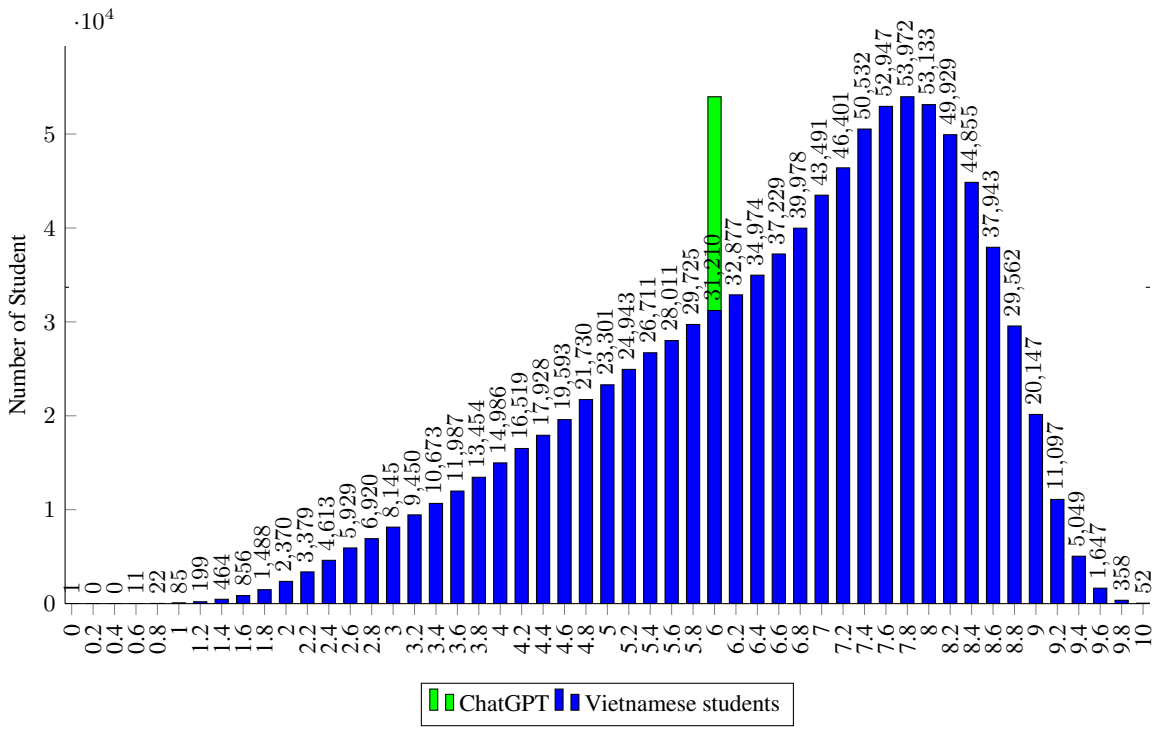

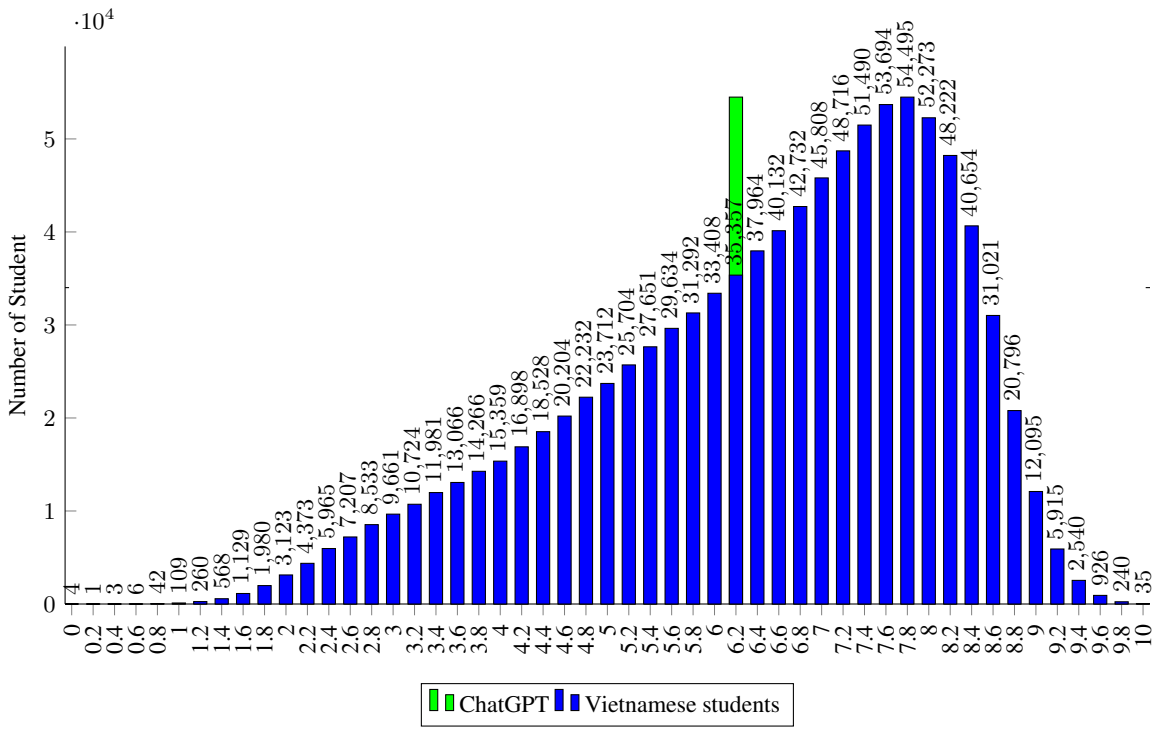
\begin{figure}[h!]
	\begin{center}	
		\begin{tikzpicture}
			\begin{axis}[
				legend style={at={(0.5,-0.125)}, 	
					anchor=north,legend columns=-1}, 
				symbolic x coords={
					0,
					0.2,
					0.4,
					0.6,
					0.8,
					1,
					1.2,
					1.4,
					1.6,
					1.8,
					2,
					2.2,
					2.4,
					2.6,
					2.8,
					3,
					3.2,
					3.4,
					3.6,
					3.8,
					4,
					4.2,
					4.4,
					4.6,
					4.8,
					5,
					5.2,
					5.4,
					5.6,
					5.8,
					6,
					6.2,
					6.4,
					6.6,
					6.8,
					7,
					7.2,
					7.4,
					7.6,
					7.8,
					8,
					8.2,
					8.4,
					8.6,
					8.8,
					9,
					9.2,
					9.4,
					9.6,
					9.8,
					10,
				},
				hide axis,
				ybar,
				bar width=5pt,
				ymin=0,
				every node near coord/.append style={rotate=90, anchor=west},
				width=\textwidth, 
				enlarge x limits={abs=0.5*\pgfplotbarwidth},
				height=10cm, 
				width=16cm,
				axis x line*=bottom, axis y line*=left
				]
				\addplot [fill=green] coordinates {
					(0,0)
				};
				\addplot [fill=blue] coordinates {
					(10,0)
				};	
				\legend{ChatGPT, Vietnamese students }	
			\end{axis}
			
			\begin{axis}[
				symbolic x coords={
					0,
					0.2,
					0.4,
					0.6,
					0.8,
					1,
					1.2,
					1.4,
					1.6,
					1.8,
					2,
					2.2,
					2.4,
					2.6,
					2.8,
					3,
					3.2,
					3.4,
					3.6,
					3.8,
					4,
					4.2,
					4.4,
					4.6,
					4.8,
					5,
					5.2,
					5.4,
					5.6,
					5.8,
					6,
					6.2,
					6.4,
					6.6,
					6.8,
					7,
					7.2,
					7.4,
					7.6,
					7.8,
					8,
					8.2,
					8.4,
					8.6,
					8.8,
					9,
					9.2,
					9.4,
					9.6,
					9.8,
					10,
				},
				hide axis,
				x tick label style={rotate=90,anchor=east},
				ybar,
				bar width=5pt,
				ymin=0,
				every node near coord/.append style={rotate=90, anchor=west},
				width=\textwidth, 
				enlarge x limits={abs=0.5*\pgfplotbarwidth},
				height=9cm, 
				width=16cm,
				axis x line*=bottom, axis y line*=left
				]
				\addplot [fill=green] coordinates {
					(0,0)
					(0.2,0)
					(0.4,0)
					(0.6,0)
					(0.8,0)
					(1,0)
					(1.2,0)
					(1.4,0)
					(1.6,0)
					(1.8,0)
					(2,0)
					(2.2,0)
					(2.4,0)
					(2.6,0)
					(2.8,0)
					(3,0)
					(3.2,0)
					(3.4,0)
					(3.6,0)
					(3.8,0)
					(4,0)
					(4.2,0)
					(4.4,0)
					(4.6,0)
					(4.8,0)
					(5,0)
					(5.2,0)
					(5.4,0)
					(5.6,0)
					(5.8,0)
					(6,0)
					(6.2,55000)
					(6.4,0)
					(6.6,0)
					(6.8,0)
					(7,0)
					(7.2,0)
					(7.4,0)
					(7.6,0)
					(7.8,0)
					(8,0)
					(8.2,0)
					(8.4,0)
					(8.6,0)
					(8.8,0)
					(9,0)
					(9.2,0)
					(9.4,0)
					(9.6,0)
					(9.8,0)
					(10,0)
					
				};	
			\end{axis}
					
			\begin{axis}[
				ylabel={Number of Student},
				legend to name={legend},
				legend style={at={(0.5,-0.175)}, 	
					anchor=north,legend columns=-1}, 
				symbolic x coords={
					0,
					0.2,
					0.4,
					0.6,
					0.8,
					1,
					1.2,
					1.4,
					1.6,
					1.8,
					2,
					2.2,
					2.4,
					2.6,
					2.8,
					3,
					3.2,
					3.4,
					3.6,
					3.8,
					4,
					4.2,
					4.4,
					4.6,
					4.8,
					5,
					5.2,
					5.4,
					5.6,
					5.8,
					6,
					6.2,
					6.4,
					6.6,
					6.8,
					7,
					7.2,
					7.4,
					7.6,
					7.8,
					8,
					8.2,
					8.4,
					8.6,
					8.8,
					9,
					9.2,
					9.4,
					9.6,
					9.8,
					10,
				},
				xtick=data,
				x tick label style={rotate=90,anchor=east},
				ybar,
				bar width=5pt,
				ymin=0,
				nodes near coords,   
				every node near coord/.append style={rotate=90, anchor=west},
				width=\textwidth, 
				enlarge x limits={abs=0.5*\pgfplotbarwidth},
				height=9cm, 
				width=16cm,
				axis x line*=bottom, axis y line*=left
				]
				\addplot [fill=blue] coordinates {
					(0,4)
					(0.2,1)
					(0.4,3)
					(0.6,6)
					(0.8,42)
					(1,109)
					(1.2,260)
					(1.4,568)
					(1.6,1129)
					(1.8,1980)
					(2,3123)
					(2.2,4373)
					(2.4,5965)
					(2.6,7207)
					(2.8,8533)
					(3,9661)
					(3.2,10724)
					(3.4,11981)
					(3.6,13066)
					(3.8,14266)
					(4,15359)
					(4.2,16898)
					(4.4,18528)
					(4.6,20204)
					(4.8,22232)
					(5,23712)
					(5.2,25704)
					(5.4,27651)
					(5.6,29634)
					(5.8,31292)
					(6,33408)
					(6.2,35357)
					(6.4,37964)
					(6.6,40132)
					(6.8,42732)
					(7,45808)
					(7.2,48716)
					(7.4,51490)
					(7.6,53694)
					(7.8,54495)
					(8,52273)
					(8.2,48222)
					(8.4,40654)
					(8.6,31021)
					(8.8,20796)
					(9,12095)
					(9.2,5915)
					(9.4,2540)
					(9.6,926)
					(9.8,240)
					(10,35)			
				};		
			\end{axis}
		\end{tikzpicture}
	\end{center}
	\caption{Mathematics score spectrum of Vietnamese students in 2022.}
	\label{fig:math_2022}
\end{figure}

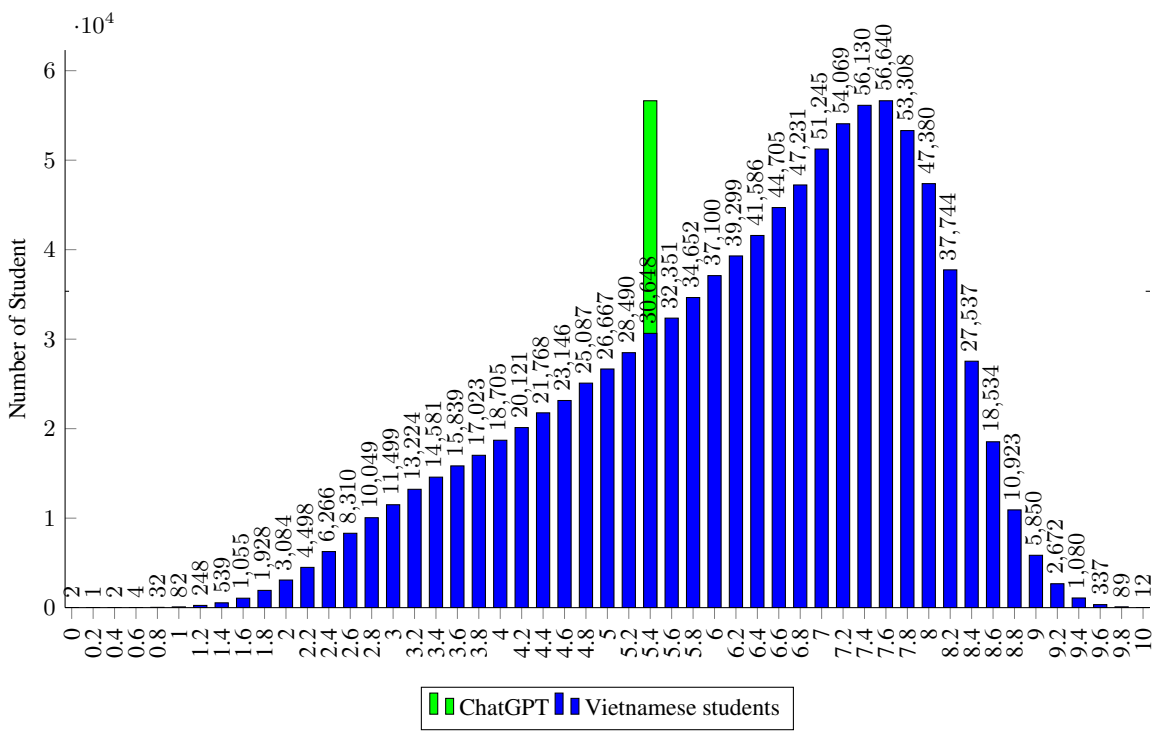
\begin{figure}[h!]
	\begin{center}	
		\begin{tikzpicture}
			\begin{axis}[
				legend style={at={(0.5,-0.125)}, 	
					anchor=north,legend columns=-1}, 
				symbolic x coords={
					0,
					0.2,
					0.4,
					0.6,
					0.8,
					1,
					1.2,
					1.4,
					1.6,
					1.8,
					2,
					2.2,
					2.4,
					2.6,
					2.8,
					3,
					3.2,
					3.4,
					3.6,
					3.8,
					4,
					4.2,
					4.4,
					4.6,
					4.8,
					5,
					5.2,
					5.4,
					5.6,
					5.8,
					6,
					6.2,
					6.4,
					6.6,
					6.8,
					7,
					7.2,
					7.4,
					7.6,
					7.8,
					8,
					8.2,
					8.4,
					8.6,
					8.8,
					9,
					9.2,
					9.4,
					9.6,
					9.8,
					10,
				},
				hide axis,
				ybar,
				bar width=5pt,
				ymin=0,
				every node near coord/.append style={rotate=90, anchor=west},
				width=\textwidth, 
				enlarge x limits={abs=0.5*\pgfplotbarwidth},
				height=10cm, 
				width=16cm,
				axis x line*=bottom, axis y line*=left
				]
				\addplot [fill=green] coordinates {
					(0,0)
				};
				\addplot [fill=blue] coordinates {
					(10,0)
				};	
				\legend{ChatGPT, Vietnamese students }	
			\end{axis}
			
			\begin{axis}[
				symbolic x coords={
					0,
					0.2,
					0.4,
					0.6,
					0.8,
					1,
					1.2,
					1.4,
					1.6,
					1.8,
					2,
					2.2,
					2.4,
					2.6,
					2.8,
					3,
					3.2,
					3.4,
					3.6,
					3.8,
					4,
					4.2,
					4.4,
					4.6,
					4.8,
					5,
					5.2,
					5.4,
					5.6,
					5.8,
					6,
					6.2,
					6.4,
					6.6,
					6.8,
					7,
					7.2,
					7.4,
					7.6,
					7.8,
					8,
					8.2,
					8.4,
					8.6,
					8.8,
					9,
					9.2,
					9.4,
					9.6,
					9.8,
					10,
				},
				hide axis,
				x tick label style={rotate=90,anchor=east},
				ybar,
				bar width=5pt,
				ymin=0,
				every node near coord/.append style={rotate=90, anchor=west},
				width=\textwidth, 
				enlarge x limits={abs=0.5*\pgfplotbarwidth},
				height=9cm, 
				width=16cm,
				axis x line*=bottom, axis y line*=left
				]
				\addplot [fill=green] coordinates {
					(0,0)
					(0.2,0)
					(0.4,0)
					(0.6,0)
					(0.8,0)
					(1,0)
					(1.2,0)
					(1.4,0)
					(1.6,0)
					(1.8,0)
					(2,0)
					(2.2,0)
					(2.4,0)
					(2.6,0)
					(2.8,0)
					(3,0)
					(3.2,0)
					(3.4,0)
					(3.6,0)
					(3.8,0)
					(4,0)
					(4.2,0)
					(4.4,0)
					(4.6,0)
					(4.8,0)
					(5,0)
					(5.2,0)
					(5.4,55000)
					(5.6,0)
					(5.8,0)
					(6,0)
					(6.2,0)
					(6.4,0)
					(6.6,0)
					(6.8,0)
					(7,0)
					(7.2,0)
					(7.4,0)
					(7.6,0)
					(7.8,0)
					(8,0)
					(8.2,0)
					(8.4,0)
					(8.6,0)
					(8.8,0)
					(9,0)
					(9.2,0)
					(9.4,0)
					(9.6,0)
					(9.8,0)
					(10,0)
					
				};	
			\end{axis}
			
			\begin{axis}[
				ylabel={Number of Student},
				legend to name={legend},
				legend style={at={(0.5,-0.175)}, 	
					anchor=north,legend columns=-1}, 
				symbolic x coords={
					0,
					0.2,
					0.4,
					0.6,
					0.8,
					1,
					1.2,
					1.4,
					1.6,
					1.8,
					2,
					2.2,
					2.4,
					2.6,
					2.8,
					3,
					3.2,
					3.4,
					3.6,
					3.8,
					4,
					4.2,
					4.4,
					4.6,
					4.8,
					5,
					5.2,
					5.4,
					5.6,
					5.8,
					6,
					6.2,
					6.4,
					6.6,
					6.8,
					7,
					7.2,
					7.4,
					7.6,
					7.8,
					8,
					8.2,
					8.4,
					8.6,
					8.8,
					9,
					9.2,
					9.4,
					9.6,
					9.8,
					10,
				},
				xtick=data,
				x tick label style={rotate=90,anchor=east},
				ybar,
				bar width=5pt,
				ymin=0,
				nodes near coords,   
				every node near coord/.append style={rotate=90, anchor=west},
				width=\textwidth, 
				enlarge x limits={abs=0.5*\pgfplotbarwidth},
				height=9cm, 
				width=16cm,
				axis x line*=bottom, axis y line*=left
				]
				\addplot [fill=blue] coordinates {
					(0,2)
					(0.2,1)
					(0.4,2)
					(0.6,4)
					(0.8,32)
					(1,82)
					(1.2,248)
					(1.4,539)
					(1.6,1055)
					(1.8,1928)
					(2,3084)
					(2.2,4498)
					(2.4,6266)
					(2.6,8310)
					(2.8,10049)
					(3,11499)
					(3.2,13224)
					(3.4,14581)
					(3.6,15839)
					(3.8,17023)
					(4,18705)
					(4.2,20121)
					(4.4,21768)
					(4.6,23146)
					(4.8,25087)
					(5,26667)
					(5.2,28490)
					(5.4,30648)
					(5.6,32351)
					(5.8,34652)
					(6,37100)
					(6.2,39299)
					(6.4,41586)
					(6.6,44705)
					(6.8,47231)
					(7,51245)
					(7.2,54069)
					(7.4,56130)
					(7.6,56640)
					(7.8,53308)
					(8,47380)
					(8.2,37744)
					(8.4,27537)
					(8.6,18534)
					(8.8,10923)
					(9,5850)
					(9.2,2672)
					(9.4,1080)
					(9.6,337)
					(9.8,89)
					(10,12)
				};		
			\end{axis}
		\end{tikzpicture}
	\end{center}
	\caption{Mathematics score spectrum of Vietnamese students in 2023.}
	\label{fig:math_2023}
\end{figure}

\section{Discussion}

While ChatGPT has certain limitations in the field of mathematics~\cite{frieder2023mathematical},\cite{azaria2022chatgpt}, \cite{borji2023categorical}, it has the potential to be a beneficial resource for educators and learners in the field of education\cite{wardat2023chatgpt},~\cite{lo2023impact}. Nevertheless, ChatGPT must continue to prove its ability to earn trust. Therefore, we need to have in-depth and detailed studies of its capabilities in areas, like mathematics. The findings of this study demonstrate that ChatGPT, a big language model trained by OpenAI, is capable of solving math issues to a certain extent but still has difficulties comprehending and interpreting graphical data in test questions. Less than the typical success rate of Vietnamese students taking the same exam, ChatGPT's total success rate in the VNHSGE exam ranged from 52$\%$ to 66$\%$. This shows that ChatGPT's capacity to tackle mathematical issues still needs to be enhanced. 

Further examination of ChatGPT's performance in resolving mathematical problems revealed that its success rate varied based on the level of difficulty and topic of the problems. The questions at the K-level had the greatest ChatGPT success rate, indicating a fundamental comprehension of the topic in question. However, the ChatGPT success rate significantly decreased as the question difficulty increased. This shows that ChatGPT has trouble solving more difficult math problems, particularly those that are at the H-level. Additionally, ChatGPT's performance varied depending on the topic. This conclusion suggests that ChatGPT's current iteration has limits in its capacity to understand mathematical ideas that call for the use of visual reasoning or the interpretation of graphical data. Future development should focus on ChatGPT's shortcomings in comprehending graphical information in test questions. This constraint could be overcome by creating algorithms and models that enable ChatGPT to read and evaluate visual data, which is crucial for resolving many mathematical issues. In summary, ChatGPT performs inconsistently across various topics and difficulty levels, although showing promising results when solving mathematical inquiries. ChatGPT's comprehension of intricate mathematical ideas, particularly those using graphical data, requires more refinement.

In our study, we compared how well ChatGPT performed in a number of well-known math competitions, including SAT Math, VNHSGE mathematics, AP Statistics, GRE Quantitative, AMC 10, AMC 12, and AP Calculus BC. The degree of difficulty, the format, and the nature of the questions employed in these contests all differ. With a 70$\%$ success rate, ChatGPT had the highest success rate in the SAT Math competition, which is not surprising considering that the SAT Math test primarily evaluates high school math proficiency. The ChatGPT success rate for the VNHSGE Mathematics, on the other hand, was 58.8$\%$. It is a more thorough test that covers a wider range of math topics and difficulty levels. It is important to note that, as was mentioned in our earlier investigation, ChatGPT performed better in some areas than others. With success rates of 25$\%$ and 1$\%$, respectively, in the GRE Quantitative and AP Calculus BC competitions, ChatGPT performed much worse. These contests are renowned for their high degree of complexity and difficulty, with questions that call for highly developed problem-solving abilities and a thorough comprehension of mathematical ideas. These types of challenges are difficult for ChatGPT to understand and analyze, which underlines the shortcomings of current language models. Overall, our analysis of ChatGPT's performance in several math competitions reveals the advantages and disadvantages of the present language models for math problem-solving. Even though language models like ChatGPT have advanced significantly in recent years, they still have difficulties processing graphical data, comprehending intricate mathematical ideas, and working out difficult mathematical problem. The goal of future study could be to overcome these constraints and improve language models' capacity for mathematical problem solving.

\section{Conclusion}

In this study, we assessed how well ChatGPT performed when it came to answering mathematics issues of various levels and topics. The findings revealed that ChatGPT performed poorly in some topics and levels while performing well in others. At Level K, ChatGPT correctly answered 83$\%$ of the questions, whereas at Levels C, A, and H, the accuracy rate dropped to 62$\%$, 27$\%$, and 10$\%$, respectively. 

Additionally, the accuracy rates of ChatGPT varied depending on the topic, with M11B, M12B, M11A, and M12D having the highest rates and M12A, M11C, and M12G having the lowest rates. It's crucial to highlight that ChatGPT had difficulty with issues requiring graphical interpretation because it couldn't read and comprehend the images, which led to a poor accuracy rate for queries about derivatives and applications.  

Furthermore, ChatGPT math scores were consistently lower than those of Vietnamese students in the same years. This might be as a result of the language model's reliance on pre-existing data and algorithms, as well as its failure to comprehend the context and nuances of the Vietnamese language.

In conclusion, ChatGPT had potential in resolving mathematical issues, but its effectiveness was constrained by elements like graphical interpretation and language understanding. Future studies might concentrate on addressing these limitations and investigating the possibilities of language models in math education.

\bibliographystyle{unsrt}  
\bibliography{paper_7_9} 

\begin{thebibliography}{10}

\bibitem{he2019practical}
Jianxing He, Sally~L Baxter, Jie Xu, Jiming Xu, Xingtao Zhou, and Kang Zhang.
\newblock The practical implementation of artificial intelligence technologies
  in medicine.
\newblock {\em Nature medicine}, 25(1):30--36, 2019.

\bibitem{chen2020artificial}
Lijia Chen, Pingping Chen, and Zhijian Lin.
\newblock Artificial intelligence in education: A review.
\newblock {\em Ieee Access}, 8:75264--75278, 2020.

\bibitem{cope2021artificial}
Bill Cope, Mary Kalantzis, and Duane Searsmith.
\newblock Artificial intelligence for education: Knowledge and its assessment
  in ai-enabled learning ecologies.
\newblock {\em Educational Philosophy and Theory}, 53(12):1229--1245, 2021.

\bibitem{Dao2021}
Xuan-Quy Dao, Ngoc-Bich Le, and Thi-My-Thanh Nguyen.
\newblock Ai-powered moocs: Video lecture generation.
\newblock In {\em 2021 3rd International Conference on Image, Video and Signal
  Processing}, pages 95--102, 2021.

\bibitem{Nguyen2021}
Thi-My-Thanh Nguyen, Thanh-Hai Diep, Bac-Bien Ngo, Ngoc-Bich Le, and Xuan-Quy
  Dao.
\newblock Design of online learning platform with vietnamese virtual assistant.
\newblock In {\em 2021 6th International Conference on Intelligent Information
  Technology}, pages 51--57, 2021.

\bibitem{vaishya2020artificial}
Raju Vaishya, Mohd Javaid, Ibrahim~Haleem Khan, and Abid Haleem.
\newblock Artificial intelligence (ai) applications for covid-19 pandemic.
\newblock {\em Diabetes \& Metabolic Syndrome: Clinical Research \& Reviews},
  14(4):337--339, 2020.

\bibitem{gao2020innovative}
Shanshan Gao.
\newblock Innovative teaching of integration of artificial intelligence and
  university mathematics in big data environment.
\newblock In {\em IOP Conference Series: Materials Science and Engineering},
  volume 750, page 012137. IOP Publishing, 2020.

\bibitem{popenici2017exploring}
Stefan~AD Popenici and Sharon Kerr.
\newblock Exploring the impact of artificial intelligence on teaching and
  learning in higher education.
\newblock {\em Research and Practice in Technology Enhanced Learning},
  12(1):1--13, 2017.

\bibitem{zhang2021ai}
Ke~Zhang and Ayse~Begum Aslan.
\newblock Ai technologies for education: Recent research \& future directions.
\newblock {\em Computers and Education: Artificial Intelligence}, 2:100025,
  2021.

\bibitem{zawacki2019systematic}
Olaf Zawacki-Richter, Victoria~I Mar{\'\i}n, Melissa Bond, and Franziska
  Gouverneur.
\newblock Systematic review of research on artificial intelligence applications
  in higher education--where are the educators?
\newblock {\em International Journal of Educational Technology in Higher
  Education}, 16(1):1--27, 2019.

\bibitem{zafari2022artificial}
Mostafa Zafari, Jalal~Safari Bazargani, Abolghasem Sadeghi-Niaraki, and Soo-Mi
  Choi.
\newblock Artificial intelligence applications in k-12 education: A systematic
  literature review.
\newblock {\em IEEE Access}, 2022.

\bibitem{pedro2019artificial}
Francesc Pedro, Miguel Subosa, Axel Rivas, and Paula Valverde.
\newblock Artificial intelligence in education: Challenges and opportunities
  for sustainable development.
\newblock 2019.

\bibitem{ahmad2021artificial}
Sayed~Fayaz Ahmad, Mohd~Khairil Rahmat, Muhammad~Shujaat Mubarik,
  Muhammad~Mansoor Alam, and Syed~Irfan Hyder.
\newblock Artificial intelligence and its role in education.
\newblock {\em Sustainability}, 13(22):12902, 2021.

\bibitem{paek2021analysis}
Seungsu Paek and Namhyoung Kim.
\newblock Analysis of worldwide research trends on the impact of artificial
  intelligence in education.
\newblock {\em Sustainability}, 13(14):7941, 2021.

\bibitem{zheng2021effectiveness}
Lanqin Zheng, Jiayu Niu, Lu~Zhong, and Juliana~Fosua Gyasi.
\newblock The effectiveness of artificial intelligence on learning achievement
  and learning perception: A meta-analysis.
\newblock {\em Interactive Learning Environments}, pages 1--15, 2021.

\bibitem{gamoran2000algebra}
Adam Gamoran and Eileen~C Hannigan.
\newblock Algebra for everyone? benefits of college-preparatory mathematics for
  students with diverse abilities in early secondary school.
\newblock {\em Educational Evaluation and Policy Analysis}, 22(3):241--254,
  2000.

\bibitem{moses2002radical}
Robert~Parris Moses, Charles~E Cobb, et~al.
\newblock Radical equations: Math literacy and civil rights.
\newblock Technical report, Beacon Press, 2002.

\bibitem{bin2022artificial}
Mohamed~Zulhilmi bin Mohamed, Riyan Hidayat, Nurain~Nabilah binti Suhaizi,
  Muhamad Khairul~Hakim bin Mahmud, Siti~Nurshafikah binti Baharuddin, et~al.
\newblock Artificial intelligence in mathematics education: A systematic
  literature review.
\newblock {\em International Electronic Journal of Mathematics Education},
  17(3):em0694, 2022.

\bibitem{hwang2022examining}
Sunghwan Hwang.
\newblock Examining the effects of artificial intelligence on elementary
  students’ mathematics achievement: A meta-analysis.
\newblock {\em Sustainability}, 14(20):13185, 2022.

\bibitem{halaweh2023chatgpt}
Mohanad Halaweh.
\newblock Chatgpt in education: Strategies for responsible implementation.
\newblock 2023.

\bibitem{Zhai2023}
Xiaoming Zhai.
\newblock {ChatGPT User Experience: Implications for Education}.
\newblock {\em SSRN Electronic Journal}, 2023.

\bibitem{kasneci2023chatgpt}
Enkelejda Kasneci, Kathrin Se{\ss}ler, Stefan K{\"u}chemann, Maria Bannert,
  Daryna Dementieva, Frank Fischer, Urs Gasser, Georg Groh, Stephan
  G{\"u}nnemann, Eyke H{\"u}llermeier, et~al.
\newblock Chatgpt for good? on opportunities and challenges of large language
  models for education.
\newblock {\em Learning and Individual Differences}, 103:102274, 2023.

\bibitem{katz2023gpt}
Daniel~Martin Katz, Michael~James Bommarito, Shang Gao, and Pablo Arredondo.
\newblock Gpt-4 passes the bar exam.
\newblock {\em Available at SSRN 4389233}, 2023.

\bibitem{gilson2023does}
Aidan Gilson, Conrad~W Safranek, Thomas Huang, Vimig Socrates, Ling Chi,
  Richard~Andrew Taylor, David Chartash, et~al.
\newblock How does chatgpt perform on the united states medical licensing
  examination? the implications of large language models for medical education
  and knowledge assessment.
\newblock {\em JMIR Medical Education}, 9(1):e45312, 2023.

\bibitem{carrascochatgpt}
JP~Carrasco, E~Garc{\'\i}a, DA~S{\'a}nchez, PD~Estrella~Porter, L~De~La~Puente,
  J~Navarro, and A~Cerame.
\newblock Is" chatgpt" capable of passing the 2022 mir exam? implications of
  artificial intelligence in medical education in spain?` es capaz
  “chatgpt” de aprobar el examen mir de 2022? implicaciones de la
  inteligencia artificial en la educaci{\'o}n.

\bibitem{frieder2023mathematical}
Simon Frieder, Luca Pinchetti, Ryan-Rhys Griffiths, Tommaso Salvatori, Thomas
  Lukasiewicz, Philipp~Christian Petersen, Alexis Chevalier, and Julius Berner.
\newblock Mathematical capabilities of chatgpt.
\newblock {\em arXiv preprint arXiv:2301.13867}, 2023.

\bibitem{OpenAI_gpt_4_report}
OpenAI.
\newblock {GPT-4 Technical Report}.
\newblock {\em arXiv preprint arXiv:2303.08774}, 2023.

\bibitem{dao2023vnhsge}
Xuan-Quy Dao, Ngoc-Bich Le, The-Duy Vo, Xuan-Dung Phan, Bac-Bien Ngo, Van-Tien
  Nguyen, Thi-My-Thanh Nguyen, and Hong-Phuoc Nguyen.
\newblock Vnhsge: Vietnamese high school graduation examination dataset for
  large language models.
\newblock {\em arXiv preprint arXiv:2305.12199}, 2023.

\bibitem{azaria2022chatgpt}
Amos Azaria.
\newblock Chatgpt usage and limitations.
\newblock 2022.

\bibitem{borji2023categorical}
Ali Borji.
\newblock A categorical archive of chatgpt failures.
\newblock {\em arXiv preprint arXiv:2302.03494}, 2023.

\bibitem{wardat2023chatgpt}
Yousef Wardat, Mohammad~A Tashtoush, Rommel AlAli, and Adeeb~M Jarrah.
\newblock Chatgpt: A revolutionary tool for teaching and learning mathematics.
\newblock {\em Eurasia Journal of Mathematics, Science and Technology
  Education}, 19(7):em2286, 2023.

\bibitem{lo2023impact}
Chung~Kwan Lo.
\newblock What is the impact of chatgpt on education? a rapid review of the
  literature.
\newblock {\em Education Sciences}, 13(4):410, 2023.

\end{thebibliography}

\end{document}